\documentclass[10pt,twocolumn,letterpaper]{article}

\usepackage[pagenumbers]{cvpr} % To force page numbers, e.g. for an arXiv version
\usepackage[accsupp]{axessibility}
\usepackage{graphicx}
\usepackage{amsmath}
\usepackage{amssymb}
\usepackage{booktabs}

\DeclareMathOperator*{\argmin}{arg\,min}
\newtheorem{theorem}{Theorem}

\usepackage[pagebackref,breaklinks,colorlinks]{hyperref}

\usepackage{xcolor}

\usepackage[capitalize]{cleveref}
\crefname{section}{Sec.}{Secs.}
\Crefname{section}{Section}{Sections}
\Crefname{table}{Table}{Tables}
\crefname{table}{Tab.}{Tabs.}

\usepackage{multirow}
\usepackage{multicol}
\usepackage{algorithm}
\usepackage{algorithmicx}
\usepackage{algpseudocode}
\usepackage{caption}
\usepackage{mathtools, cuted}
\usepackage{lipsum, color}
\def\cvprPaperID{7271} % *** Enter the CVPR Paper ID here
\def\confName{CVPR}
\def\confYear{2023}

\begin{document}

\title{IDGI: A Framework to Eliminate Explanation Noise from Integrated Gradients}

\author{Ruo Yang, Binghui Wang, and Mustafa Bilgic\\
Department of Computer Science, Illinois Institute of Technology, Chicago, IL, USA\\
{\tt\small ryang23@hawk.iit.edu, bwang70@iit.edu, mbilgic@iit.edu}
}
\maketitle

%%%%%%%%% ABSTRACT
\begin{abstract}
Integrated Gradients (IG) as well as its variants are well-known techniques for interpreting the decisions of deep neural networks. While IG-based approaches attain state-of-the-art performance, they often integrate noise into their explanation saliency maps, which reduce their interpretability. To minimize the noise, we examine the source of the noise analytically and propose a new approach to reduce the explanation noise based on our analytical findings. We propose the Important Direction Gradient Integration (IDGI) framework, which can be easily incorporated into any IG-based method that uses the Reimann Integration for integrated gradient computation. Extensive experiments with three IG-based methods show that IDGI improves them drastically on numerous interpretability metrics. The source code for IDGI is available at \url{https://github.com/yangruo1226/IDGI}.
\end{abstract}

%%%%%%%%% Introduction
\section{Introduction}
\label{sec_intro}

\begin{figure}[!t]
     \centering
     \begin{subfigure}[b]{0.47\textwidth}
         \centering
         \includegraphics[width=\textwidth]{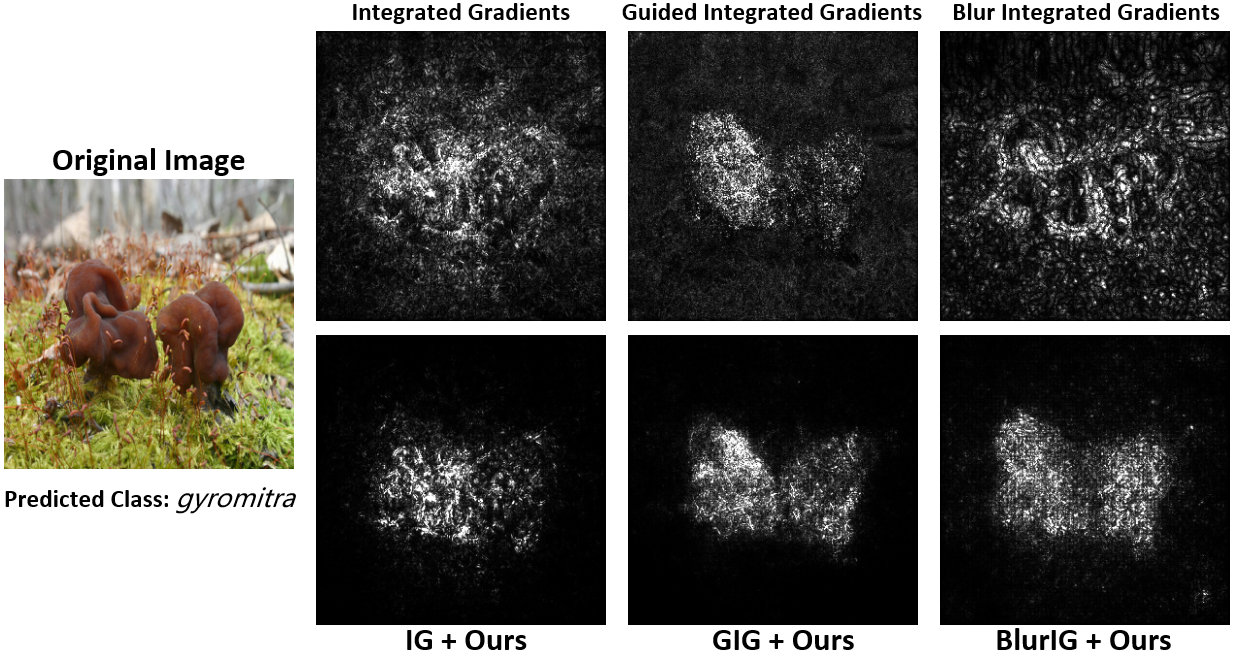}
     \end{subfigure}
\caption{Saliency/attribution map of the existing IG-based methods and those with our method on explaining the prediction from \textit{InceptionV3} model. Our method significantly reduces the noise in the saliency map created by these IG-based methods.}
\label{fig_intro}
\vspace{-6mm}
\end{figure}

With the deployment of deep neural network (DNN) models for safety-critical applications such as autonomous driving \cite{chen2017multi,Caesar_2020_CVPR,Chen_2015_ICCV} and medical diagnostics \cite{liu2021relational, cuadros2009eyepacs}, explaining the decisions of DNNs has become a critical concern. For humans to trust the decision of DNNs, not only the model must perform well on the specified task, it also must generate explanations that are easy to interpret. A series of explanation methods (e.g., gradient-based saliency/attribution map approaches \cite{simonyan2013deep, sundararajan2017axiomatic, kapishnikov2021guided, xu2020attribution, kapishnikov2019xrai, selvaraju2017grad, smilkov2017smoothgrad, pan2021explaining} as well as many that are not based on gradients \cite{lundberg2017unified, sundararajan2020many, jethani2021fastshap, vstrumbelj2014explaining, ribeiro2016should, fong2017interpretable, dabkowski2017real, zintgraf2017visualizing, petsiuk2018rise, zeiler2014visualizing, springenberg2014striving, shrikumar2017learning, bach2015pixel, montavon2019layer}) have been developed to connect a DNN's prediction to its input. Among them, Integrated Gradients (IG) \cite{sundararajan2017axiomatic}, a well-known gradient-based explanation method, and its variants \cite{kapishnikov2021guided, xu2020attribution} have attracted significant interest due to their state-of-the-art explanation performance and desirable theoretical properties. However, we observe that explanation noise exists in the attribution generated by these IG methods (please see \cref{fig_intro}). 
In this research, we investigate IG-based methods, study the explanation noise introduced by these methods, propose a framework to remove the explanation noise, and empirically validate the effectiveness of our approach.

A few recent IG-based methods (e.g., \cite{smilkov2017smoothgrad} \cite{kapishnikov2021guided},
\cite{xu2020attribution}, 
\cite{sturmfels2020visualizing})
have been proposed to address the noise issue. Kapishnikov et al.\cite{kapishnikov2021guided} provide the following main reasons\footnote{\cite{kapishnikov2021guided} mentions the accuracy of integration is also a reason to generate the noise, but this is not the focus of existing IG methods and this paper.} that could generate the  noise: 1) DNN model's shape often has a high curvature; and 2) The choice of the reference point impacts explanation. They propose Guided Integrated
Gradients (GIG)~\cite{kapishnikov2021guided}, which tackles point \#1 by iteratively finding the integration path that tries to avoid the high curvature points in the space. Blur Integral Gradients~\cite{xu2020attribution}, on the other hand, shows that the noise could be reduced by finding the integration path through the frequency domain instead of the original image domain. 
Formally, it finds the path by successively blurring the input via a Gaussian blur filter. Sturmfels et al.\cite{sturmfels2020visualizing} tackle point \#2 by performing the integration from multiple reference points, while Smilkov et al.\cite{smilkov2017smoothgrad} aggregate the attribution with respect to multiple Gaussian noisy inputs to reduce the noise. 
{Nevertheless, all IG-based methods share a common point in that they 
compute the integration of gradients via the Riemann integral.} 
\emph{We highlight that, the integration calculation by the existing methods fundamentally 
introduces the explanation noise.} To this end, we offer a general solution that eliminates the noise by examining the integration directions from the explanation perspective.

Specifically, we investigate each computation step in the Riemann Integration and then theorize about the noises' origin. Each Riemann integration calculation integrates the gradient in the \textit{original direction}---it first computes the gradient with respect to the starting point of the current path segment and then multiplies the gradient by the path segment. We show that the \textit{original direction} can be divided into an \textit{important direction} and a  \textit{noise direction}. We theoretically demonstrate that the true gradient is orthogonal to the \textit{noise direction}, resulting in the gradient's multiplication along the noise direction having no effect on the attribution. Based on this observation, we design a framework, termed Important Direction Gradient Integration (IDGI), that can eliminate the explanation noise in each step of the computation in any existing IG method.
Extensive investigations reveal that IDGI reduces noise significantly when evaluated using state-of-the-art IG-based methods.

In summary, our main contributions are as follows:
\begin{itemize}
  \item We propose the Important Direction Gradient Integration (IDGI), a general framework to eliminate the explanation noise in IG-based methods, and investigate its theoretical properties. 
  \item We propose a novel measurement for assessing the attribution techniques' quality, i.e., AIC and SIC using MS-SSIM. We show that this metric offers a more precise measurement than the original AIC and SIC.
  \item Our extensive evaluations on 11 image classifiers with 3 existing and 1 proposed attribution assessment techniques indicate that IDGI significantly improves the attribution quality over the existing IG-based methods.
\end{itemize}

%%%%%%%%% Related Work

\section{Background}

\label{sec_back}
\subsection{Integrated Gradient and its Variants}

\noindent {\bf Integrated Gradients (IG) \cite{sundararajan2017axiomatic}.}
Given a classifier $f$, class $c$, and input $x$, the output $f_c(x)$ represents the confidence score (e.g., probability) for predicting $x$ to class $c$. 
IG computes the importance/attribution per feature (e.g., a pixel in an image) by calculating the line integral between the reference point $x'$ and image $x$ in the vector field that the model creates, where the vector field is formed by the gradient of $f_{c}(x)$ with respect to the input space. Formally, for each feature $i$, the IG, is defined as below:
\begin{equation}
\label{ig_re}
        I_{i}^{IG}(x)  =\int_{0}^{1}\frac{\partial f_{c}(\gamma^{IG}(\alpha))}{\partial \gamma^{IG}_{i}(\alpha)} \frac{\partial \gamma^{IG}_{i}(\alpha)}{\partial \alpha} d \alpha, 
\end{equation}
where $\gamma^{IG}(\alpha), \alpha \in [0,1]$ is the parametric function representing the path from $x'$ to $x$, e.g.$\gamma^{IG}(0) = x', \gamma^{IG}(1) = x$. Specifically, $\gamma^{IG}$ is a straight line that connects $x'$ and $x$.

\vspace{+0.5mm}

\noindent {\bf Guided Integrated Gradients (GIG) \cite{kapishnikov2021guided}.} Kapishnikov et al.~\cite{kapishnikov2021guided} claim that DNN's output shape has a high degree of curvature; hence, the larger-magnitude gradients from each feasible point on the path would have a significantly 
larger effect on the final attribution values. To address it, they propose GIG, an adaptive path method for noise reduction. 
Specifically, instead of integrating gradients following the straight path as IG does, GIG initially seeks a new path to avoid high-gradient directions as below:
\vspace{-2mm}
\begin{equation}
    \gamma^{GIG} = \argmin_{\gamma \in \Gamma} \sum_{i=1}^N \int_{0}^{1} | \frac{\partial f_{c}(\gamma(\alpha))}{\partial \gamma_{i}(\alpha)} \frac{\partial \gamma_{i}(\alpha)}{\partial \alpha} | d \alpha,
\end{equation}
\noindent where $\Gamma$ is the set containing all possible path between $x^{'}$ and $x$. After finding the optimal path $\gamma^{GIG}$, GIG uses it and computes the attribution values similar to IG. Formally, 
\begin{equation}
\label{gig_re}
        I_{i}^{GIG}(x)  =\int_{0}^{1}\frac{\partial f_{c}(\gamma^{GIG}(\alpha))}{\partial \gamma^{GIG}_{i}(\alpha)} \frac{\partial \gamma^{GIG}_{i}(\alpha)}{\partial \alpha} d \alpha.
\end{equation}

\noindent {\bf Blur Integrated Gradients (BlurIG) \cite{xu2020attribution}.} Xu et al.~\cite{xu2020attribution} propose BlurIG, which integrates the IG into frequency domain as opposed to the original image domain. In other words, BlurIG takes into account the path produced by sequentially blurring the input with a Gaussian blur filter. Specifically, suppose the image has the 2D shape $M \times N$, and let $x(p,q)$ represent the value of the image $x$ at the location of pixels $p$ and $q$. The discrete convolution of the input signal with a 2D Gaussian kernel with variance $\alpha$ is thus defined as follows:
\begin{align}
    \gamma^{BlurIG}&::=L(x,p,q,\alpha) \nonumber \\ &= \sum_{m=-\infty}^{\infty}\sum_{n=-\infty}^{\infty} \frac{1}{\pi\alpha} e^{-\frac{q^2+p^2}{\alpha}}x(p-m,q-n)
\end{align}

\noindent Then, the final BlurIG computation is as below:
\begin{equation}
\label{blurig_re}
        I_{p,q}^{BlurIG}(x)  =\int_{\infty}^{0}\frac{\partial f_{c}(\gamma^{BlurIG}(\alpha))}{\partial \gamma^{BlurIG}_{p,q}(\alpha)} \frac{\partial \gamma^{BlurIG}_{p,q}(\alpha)}{\partial \alpha} d \alpha, 
\end{equation}

\subsection{Riemann Integration}

Existing IG-based methods (IG: \cref{ig_re}, GIG: \cref{gig_re}, and BlurIG: \cref{blurig_re}) all approximate the line integral numerically using the Riemann Integration. Specifically, as shown in \cref{eq2}, they discretize the path between $x{'}$ and $x$ into $N$, a large and finite number, small piece-wise linear segments, and aggregate the value of multiplication between the gradient vectors and the small segments:
\begin{align}
\label{eq2}
    & f_c(x) - f_c(x{'})  = \lim_{N \rightarrow \infty} \sum_{j=0}^{N} \frac{\partial f_c(x_j)}{\partial x_j}{(x_{j+1} - x_j)} \nonumber \\
    & \qquad = \sum_{i=0}^{n}I_{i}(x)  =\int_{0}^{1}\frac{\partial f_{c}(\gamma(\alpha))}{\partial \gamma(\alpha)} \frac{\partial \gamma(\alpha)}{\partial \alpha} d \alpha 
\end{align}

\noindent In other words, regardless of whichever approach (IG, GIG, or BlurIG) is used for calculating the attributions, the final attribution map is produced from Riemann Integration in all IG-based algorithms. As the algorithm approximates the integration discretely, the approximation itself contains numerical inaccuracies relative to the theoretical real values. However, we do not concentrate on eliminating numerical errors; rather, we discovered that each Riemann Integration step creates noise from an explanation perspective (please see \cref{illustration_fig}). Specifically, each path segment has a \textit{noise direction} where gradient integration with that direction contributes nothing to output values, which indicates the attribution values generated with this direction are noisy.

%%%%%%%%% Approach
\section{Our Framework: Important Direction Gradient Integration (IDGI)}
\label{sec_app}

In this section, we first describe the concept of \textit{important direction} and the noise that arises from the gradient with the \textit{noise direction} to the attributions. Then, we formally introduce the Important Direction Gradient Integration (IDGI), a framework that only leverages the gradient with the \textit{important direction}. We highlight that IDGI can be applied to any IG-based method. Finally, we discuss the theoretical properties of IDGI.

\subsection{Important Direction and Noise Direction}
\vspace{-2mm}
\label{iddac}
\begin{figure}[!t]
     \centering
     \begin{subfigure}[b]{0.48\textwidth}
         \centering
         \includegraphics[width=\textwidth]{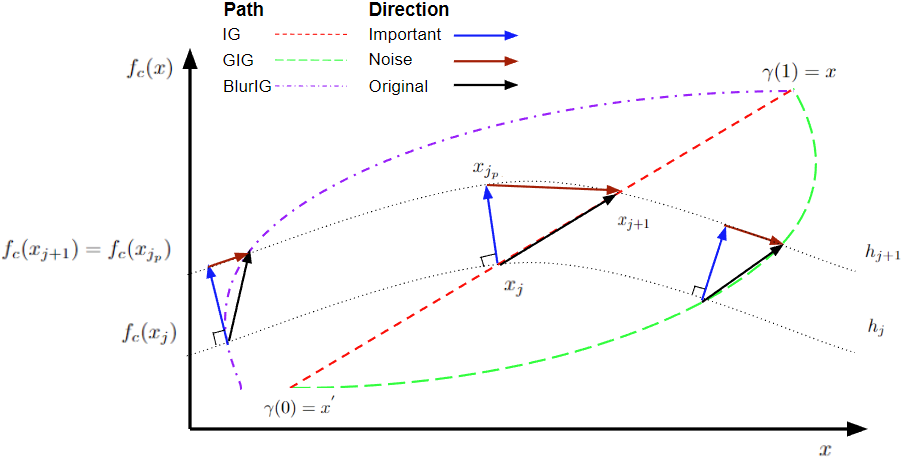}
     \end{subfigure}
\caption{Illustration of IDGI. The {\color{red} red}, {\color{green} green}, and {\color{violet} purple} lines illustrate the path corresponding to IG, GIG, and BlurIG, respectively, where both IG and GIG need a reference point as manual input but BlurIG does not. Riemann Integration computes the integral by multiplying the {\color{blue} blue} and {\bf black} vectors at each calculation step $j$. However, each {\bf black} vector can be linearly decomposed of {\color{blue} blue} and {\color{brown} brown} vectors. Since the direction of the {\color{blue} blue} vector, i.e. gradient, represents the maximum rate of change for the function value, we consider it to be the most \textit{important direction}. Instead of multiplying the {\color{blue} blue} vector by the {\bf black} vector, we propose computing the integral using the {\color{blue} blue} vector alone. In addition to the fact that the multiplication of the {\color{blue} blue} and {\color{brown} brown} vectors has no effect on the function's value, we assert that this integration also creates the noise.}
\label{illustration_fig}
\vspace{-4mm}
\end{figure}

\noindent {\bf Important Direction.} Given the point $x_j = \gamma(\alpha_j)$ and its next point $x_{j+1} = \gamma(\alpha_{j+1})$ on the path from reference point $x{'}$ to the input point $x$, IG-based methods compute the gradient, $g$, of $f_c(x_j)$ with respect to $x$ and utilize Riemann integration to multiply element-wisely with the direction (\textit{original direction}) from $x_j$ to $x_{j+1}$ (please see \cref{illustration_fig}). Based on the Riemann integration principle, when $N$ increases, the sum of the multiplication result accurately estimates the differences in the function values, $f_c(x_{j+1})-f_c(x_j)$. In terms of the explanation, the multiplication result for this step indicates the contribution to change in the value of the function from $f_c(x_j)$ to $f_c(x_{j+1})$. In other words, attribution values at this step explain why the prediction score moves from $f_c(x_j)$ to $f_c(x_{j+1})$.

The unit direction vector of the gradient at step $j$, i.e., $\frac{g}{|g|}$, indicates the direction of the fastest increase of the function $f_c$ at $x_j$. That is, moving the point $x_j$ along the direction $g$ changes $f_c$ the most. We refer to the direction $\frac{g}{|g|}$ as the \textit{important direction}. In general, the gradient of the function value $f_c$ at each point in space defines the conservative vector field, where an infinite number of hyperplanes $h$ exist, and each hyperplane contains all points with the same function output value. For instance, $x_j$ resides on the hyperplane $h_j$ if all the points $x_{h_j} \in h_j$ have the same function value $f_c(x_j)$, i.e., $f_c(x_{h_j}) = f_c(x_{j}), \forall x_{h_j} \in h_j$. In the conservative vector field, separate hyperplanes never intersect, which means that each point has its own projection point with regard to the other hyperplanes. Moreover, to identify the projections, one may move the point from its original hyperplane toward the next hyperplane in which the moving direction is the same as the gradient's direction $\frac{g}{|g|}$. Similarly, $x_{j+1}$ stays on the hyperplane $h_{j+1}$ where $f_c(x_{h_{j+1}}) = f_c(x_{j+1}), \forall x_{h_{j+1}} \in h_{j+1}$. For point $x_j$, if one moves $x_j$ along the \textit{important direction}, there exists an unique projection point $x_{j_{p}}$ on the hyperplane $h_{j+1}$ where $f_c(x_{j_{p}}) = f_c(x_{j+1})$. As we illustrate in \cref{theorem_1}, while the attribution for each feature $i$ computed from the \textit{original direction} ($x_{j+1} - x_{j}$) and \textit{important direction} ($x_{j_{p}} - x_j$) could be different, the change in the value of the function $f_c$, which are the prediction values to be explained are the same since $x_{j+1}$ and $x_{j_{p}}$ are on the same hyperplane $h_{j+1}$.

\begin{theorem}
\label{theorem_1}
Given a function $f_c(x): R^n \rightarrow R$, points $x_j, x_{j+1}, x_{j_p} \in R^n$, then the gradient of the function with respect to each point in the space $R^n$ forms the conservative vector fields $\overrightarrow{F}$ and further define the hyperplane $h_j=\{x: f_c(x)=f_c(x_j)\}$ in $\overrightarrow{F}$. Assume the Riemann Integration accurately estimates the line integral of the vector field $\overrightarrow{F}$ from points $x_j$ to $x_{j+1}$ and $x_{j_p}$ e.g. $\int_{x_j}^{x_{j_p}}\frac{\partial f_c(x)}{\partial x} dx \approx \frac{\partial f_c(x_j)}{\partial x_j}{(x_{j_p} - x_j)}$, and $x_j \in h_j$, $x_{j_p}, x_{j+1} \in h_{j+1}$. Then:
\[\int_{x_j}^{x_{j+1}}\frac{\partial f_c(x)}{\partial x} dx \approx \int_{x_j}^{x_{j_p}}\frac{\partial f_c(x)}{\partial x} dx.\]
\end{theorem}

\noindent {\bf Noise Direction.} For step $j$, any \textit{original direction} can be decomposed into a combination of \textit{important direction} and \textit{noise direction}. Therefore, the integration at step $j$ consists of two parts: integration from $x_j$ to $x_{j_{p}}$ and from $x_{j_{p}}$ to $x_{j+1}$. While (i) integrating from $x_j$ to $x_{j_p}$ (\textit{important direction}) and then $x_{j_p}$ to $x_{j+1}$ (\textit{noise direction}) and (ii) integrating directly from $x_j$ to $x_{j+1}$ (\textit{original direction}) often assign different attribution values to the features, the target predicted score $f_c(x_{j+1})-f_c(x_j)$ to be explained at step $j$ stays the same regardless of which path is chosen.

Since $f_c(x_{j_p}) \approx f_c(x_{j+1})$ (as we demonstrated in \cref{theorem_1}), the integration of the first path, i.e., from $x_j$ to $x_{j_{p}}$, of the two-parts-path offers the full attributions that explain the prediction value changes from $f_c(x_j)$ to $f_c(x_{j+1})$. This suggests that the second path attributions (from $x_{j_p}$ to $x_{j+1}$) do not account for any changes in prediction score values. We refer to this direction of the second path as the \textit{noise direction} and argue that the integration with this direction adds noise to the attribution since it explains zero contribution for changes in the prediction score.

\begin{algorithm}[!t]
    \caption{Important Direction Gradient Integration}
    \label{idgi_algo}
    \begin{algorithmic}[1]
    \Statex Inputs: $x$, $f$, $c$, $path: [x',\dots, x_j,\dots, x]$
    \State $I_{i}^{IDGI} = 0$
    \For{$x_j \in path$}
        \State $d = f_{c}(x_{j+1}) - f_{c}(x_{j})$
        \State $g = \frac{\partial f_c(x_j)}{\partial x}$
        \State $I_{i}^{IDGI} \mathrel{{+}{=}} \frac{g_i \times g_i \times d}{g \cdot g}$
    \EndFor
    \State \Return $I^{IDGI}$
    \end{algorithmic}
\end{algorithm}

\setlength{\textfloatsep}{+2mm}

\subsection{IDGI Algorithm}
\vspace{-2mm}
In order to reduce the noise from the attributions, we propose the Important Direction Gradient Integration (IDGI) framework. Suppose we are given an input $x$, target class $c$, classifier $f$, and a given path $path: [x',\dots, x_j,\dots, x]$ from any IG-based method. 
Similar to IG-based approaches, IDGI first calculates the gradient $g$ with respect to the current point at each Riemann integration step. IDGI then determines the unit \textit{important direction} vector of $g$ as $\frac{g}{|g|}$ and the step size $\frac{f_c(x_{j+1})-f_c(x_{j})}{|g|}$ rather than multiplying $g$ with the distance $x_{j+1}-x_j$. The step size determines how much of the unit direction vector should be applied and ensures that the current step explains the change in probability from $f_c(x_{j+1})-f_c(x_{j})$. In other words, the projection of $x_j$ onto the hyperplane $h_{j+1}$, formed as $x_{j_p} = x_j+ \frac{g}{|g|}\cdot\frac{f_c(x_{j+1})-f_c(x_{j})}{|g|}$, has the same function value as point $x_{j+1}$, i.e., $f_c(x_{j+1})=f_c(x_{j_p})$, since both $x_{j+1}$ and $x_{j_p}$ reside on the hyperplane $h_{j+1}$. Finally, IDGI aggregates the computations from each step and forms the final attributions to interpret the prediction score of $f_c(x)$. The pseudo-code of IDGI is described in \cref{idgi_algo}.

\subsection{Theoretical Properties of IDGI}
\label{properties_idgi}
Sundararajan et al. \cite{sundararajan2017axiomatic} introduce several axioms that any explanation method is expected to adhere to. These axioms include Completeness, Sensitivity(a, b), Implementation Invariance, Linearity, and Symmetry preserving. As we discuss below, IDGI satisfies all of them except Linearity.

\noindent {\bf Completeness.} IDGI clearly satisfies this axiom since the computation for $j^{th}$ step in IDGI computes the attribution for $f_c(x_{j+1})-f_c(x_{j})$ exactly. This also indicates that IDGI satisfies \textit{Senstivity-N} \cite{ancona2017towards}.

\noindent {\bf Sensitivity(a).} IDGI satisfies \textit{Sensitivity(a)} at each computation step $j$ (from $x_j$ to $x_{j+1}$) if the underlying IG-based method to which IDGI is applied satisfies the axiom. Consider a case that during the $j^{th}$ computation step, the baseline at this step refers to $x_{j}$, input refers to $x_{j+1}$, and the two points differ only at $i^{th}$ feature, e.g. $x^{q}_{j} \neq x^{q}_{j+1}, \forall q = i$ and $x^{q}_{j} = x^{q}_{j+1}, \forall q \neq i$. Also, assume the predictions vary due to the $i^{th}$ feature differences, e.g. $f_c(x_{j}) \neq f_c(x_{j+1})$. If the underlying IG-based method satisfies \textit{Sensitivity(a)}, then the $i^{th}$ of the computed gradient $g$ at this step is non-zero, i.e., $g_i \neq 0$. Then IDGI assigns non-zero value to the attribution of feature $i$ at step $j$ since $g_i \times g_i \times (f_c(x_{j+1}) - f_c(x_{j})) \neq 0$. From $x{'}$ to $x$, IDGI satisfies \textit{Sensitivity(a)} if the underlying IG-based method, to which IDGI is applied, satisfies the axiom \emph{and} the function evaluation on the giving path $p$ is strictly monotonic, e.g. $f_c(x_{j+1}) > f_c(x_{j})$ or $f_c(x_{j+1}) < f_c(x_{j}), \forall x_j \in p$. Since the underlying IG-based method satisfies \textit{Sensitivity(a)}, it indicates that there exists at least one gradient $g$ which has a non-zero value for $i^{th}$ feature, i.e., $g_i \neq 0$, during all the computation steps. Then, the strictly monotonic property guarantees the non-zero value is captured in the final attribution for $i^{th}$ feature. Also, since the square value of $g_i$ is computed during each step, the strictly monotonic property assures that any attribution value added to the final attribution contributes in the same direction rather than canceling each other. Furthermore, intuitively, the strictly monotonic property of the function evaluation on the giving path $p$ is expected since the changing of one feature in one direction is expected to impact the output of the function in one direction too, e.g increasing the prediction value.

\noindent {\bf Sensitivity(b).} IDGI satisfies this axiom as long as the underlying IG-based method satisfies it. If the IG-based method satisfies \textit{Sensitivity(b)}, then any variable/feature $i$ that the function $f_c$ does not rely on will have zero gradient value everywhere, i.e., $g_i = 0$. Clearly, the final attribution of IDGI also assigns the value of zero to such a feature $i$ since $g_i$ is zero in each step of the computation.

\noindent {\bf Implementation Invariance.} Kapishnikov et al.\cite{kapishnikov2021guided} showed an attribution method that depends only on function gradients and is independent of the network's underlying structure satisfies the \textit{Implementation Invariance} axiom. Thus, IDGI preserves the invariance as long as the underlying IG-based method is independent of the network's topology and solely depends on the gradients of the functions.

\noindent {\bf Linearity.} Given a linearly combined network $f_3 = a\times f_1 + b\times f_2$, Linearity requires that the explanation method assign attribution values as a weighted combination of attribution from $f_1$ and $f_2$, i.e, $\text{IG}_{f_3}(x) = a\times\text{IG}_{f_1}(x)+b\times\text{IG}_{f_2}(x)$. IDGI does not satisfy this requirement. However, in practice, we often try to explain the predictions of a complex nonlinear function, such as DNNs, instead of the linear composition of models.

\noindent {\bf Symmetry preserving.} An attribution method is \textit{Symmetry preserving} if, for all inputs that have identical values for symmetric variables and baselines that have identical values for symmetric variables, the symmetric variables receive identical attributions \cite{sundararajan2017axiomatic}. IDGI satisfies this axiom only if the underlying IG-based method satisfies it.

%%%%%%%%% Experiments
\section{Experimental Methodology and Results}
\label{sec_exp}

\subsection{Experimental Setup}
\vspace{-2.0mm}
\noindent {\bf Baselines.} We use four gradient-based methods as baselines. Since the Integrated Gradients (IG) \cite{sundararajan2017axiomatic}, Guided Integrated Gradients (GIG) \cite{kapishnikov2021guided}, and Blur Integrated Gradients (BlurIG) \cite{xu2020attribution} offer distinct paths from the reference point to the image of interest, our approach could be applied independently to any of them because it is orthogonal to any given path. Additionally, we also include the Vanilla Gradient (VG)\cite{simonyan2013deep}, which takes the gradient of the output $f_c(c)$ with respect to $x$. We use the original implementations\footnote{https://github.com/PAIR-code/saliency} with default parameters in the authors' code for IG, GIG, and BlurIG, and implement our method as an additional module that interfaces with the original implementations. Then, same as \cite{kapishnikov2021guided}, we use a step size of 200 as the parameter needed to compute the Riemann integral for all methods except VG since it doesn't use Riemann Integration. As is the common practice, we use the black image as the reference point for IG and GIG.

\noindent {\bf Models.} We use the TensorFlow (2.3.0)\cite{tensorflow2015-whitepaper} pre-trained models: DenseNet\{121, 169, 201\} \cite{huang2017densely}, InceptionV3 \cite{szegedy2016rethinking}, MobileNetV2 \cite{howard2017mobilenets}, ResNet\{50, 101, 151\}V2 \cite{he2016deep}, VGG\{16, 19\} \cite{simonyan2014very}, and Xception \cite{chollet2017xception}.

\noindent {\bf Dataset.} Following the research \cite{kapishnikov2021guided, xu2020attribution, kapishnikov2019xrai, pan2021explaining}, we use the Imagenet \cite{deng2009imagenet} validation dataset, which contains 50K test samples with labels and annotations. We test the explanation methods for each model on images that the model predicted the label correctly. Hence, the number of test images varies from 33K to 39K corresponding to different models.

\noindent {\bf Evaluation metrics.} We use numerous metrics to quantitatively evaluate our method and compare it with baselines: Insertion score \cite{pan2021explaining, petsiuk2018rise}, the Softmax information curves (SIC), the Accuracy information curves (AIC) \cite{kapishnikov2019xrai, kapishnikov2021guided}, and three Weakly Supervised Localization\cite{cong2018review, xu2020attribution, kapishnikov2019xrai} metrics. We also introduce a modified version of SIC and AIC with MS-SSIM information level. We implement all the evaluation metrics as they were introduced in the previous works, and we discuss the implementation details in \cref{exp_insertion}-\ref{exp_f1rocmae}.

\subsection{Qualitative Check}

\begin{figure}[!t]
     \centering
     \begin{subfigure}[b]{.48\textwidth}
         \centering
         \includegraphics[width=\textwidth]{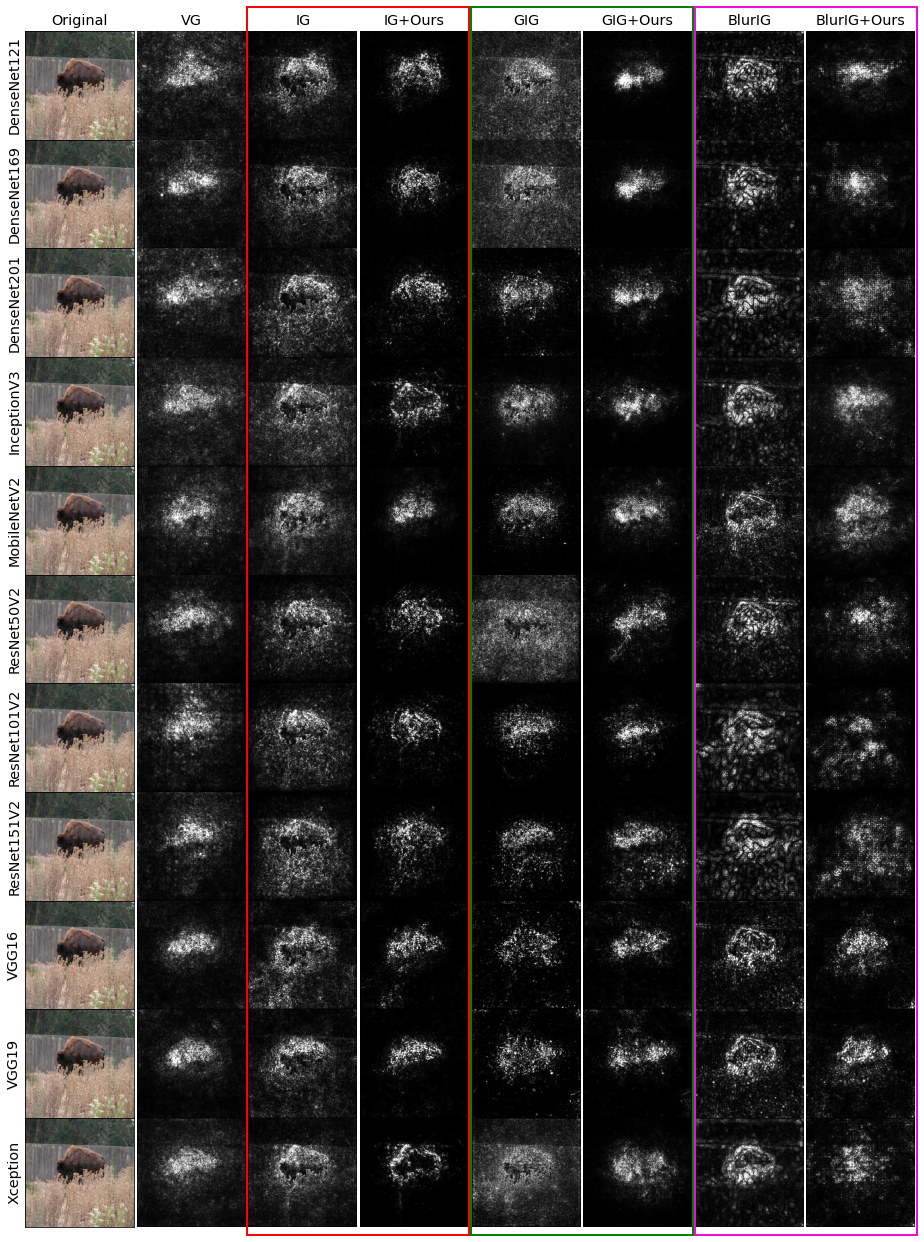}
     \end{subfigure}
     \vspace{-4mm}
\caption{Saliency map comparisons between different methods and different models (each row). Each model correctly predicts the image as \textit{bison}. Left-to-right: Original image, VG, IG, IG$+$IDGI, GIG, GIG$+$IDGI, BlurIG, and BlurIG$+$IDGI. Methods with IDGI focus more on the region of the \textit{bison}.
}
\label{fig_exp0}
\end{figure}

\cref{fig_exp0} shows a sample of observations for all baselines and IDGI with all models. Compared to baseline methods, the outcome of our method demonstrates the relevant pixels are more concentrated in the bison region of the original image. Also, the noises are less in other regions. However, qualitative and visual inspections are often subjective and hence we focus more on the quantitative metrics in the rest of the experiments.

\begin{table}[!t]
\centering
\resizebox{\columnwidth}{!}{
\begin{tabular}{|c|c||c|c||c|c||c|c||c|}
\hline
\textbf{Metrics}&\textbf{Models}&\multicolumn{6}{|c|}{\textbf{IG-based Methods}}&{\textbf{Other}}\\
\cline{3-9}
&&IG&+Ours&GIG&+Ours&BlurIG&+Ours&VG\\
\hline

\cline{3-9}
\hline
\cline{3-9}
\hline
\cline{3-9}
\hline
\cline{3-9}
\hline
\multirow{8}{*}{\rotatebox[origin=c]{0}{\shortstack{Insertion \\ Score \\ with \\ Probability \\($\uparrow$)}}}&\textit{DenseNet121}&.293&\textbf{.441}&.304&\textbf{.364}&.250&\textbf{.357}&.176\\
&\textit{DenseNet169}&.350&\textbf{.503}&.362&\textbf{.421}&.298&\textbf{.412}&.203\\
&\textit{DenseNet201}&.329&\textbf{.457}&.341&\textbf{.397}&.283&\textbf{.395}&.204\\
&\textit{InceptionV3}&.348&\textbf{.478}&.368&\textbf{.460}&.305&\textbf{.446}&.214\\
&\textit{MobileNetV2}&.203&\textbf{.346}&.231&\textbf{.291}&.198&\textbf{.306}&.129\\
&\textit{ResNet50V2}&.338&\textbf{.431}&.364&\textbf{.403}&.256&\textbf{.380}&.219\\
&\textit{ResNet101V2}&.377&\textbf{.459}&.386&\textbf{.412}&.290&\textbf{.400}&.248\\
&\textit{ResNet151V2}&.388&\textbf{.480}&.388&\textbf{.418}&.294&\textbf{.411}&.242\\
&\textit{VGG16}&.217&\textbf{.338}&.212&\textbf{.265}&.192&\textbf{.298}&.134\\
&\textit{VGG19}&.232&\textbf{.342}&.224&\textbf{.273}&.206&\textbf{.306}&.145\\
&\textit{Xception}&.334&\textbf{.487}&.367&\textbf{.475}&.298&\textbf{.453}&.215\\

\hline
\hline

\multirow{8}{*}{\rotatebox[origin=c]{0}{\shortstack{Insertion \\ Score \\ with \\ Probability \\ Ratio \\($\uparrow$)}}}&\textit{DenseNet121}&.322&\textbf{.484}&.337&\textbf{.400}&.276&\textbf{.392}&.194\\
&\textit{DenseNet169}&.360&\textbf{.518}&.374&\textbf{.434}&.307&\textbf{.424}&.210\\
&\textit{DenseNet201}&.354&\textbf{.492}&.370&\textbf{.428}&.307&\textbf{.425}&.222\\
&\textit{InceptionV3}&.389&\textbf{.532}&.413&\textbf{.512}&.342&\textbf{.496}&.240\\
&\textit{MobileNetV2}&.248&\textbf{.419}&.285&\textbf{.354}&.244&\textbf{.369}&.160\\
&\textit{ResNet50V2}&.353&\textbf{.450}&.382&\textbf{.422}&.268&\textbf{.397}&.229\\
&\textit{ResNet101V2}&.389&\textbf{.473}&.399&\textbf{.425}&.300&\textbf{.412}&.256\\
&\textit{ResNet151V2}&.399&\textbf{.494}&.400&\textbf{.430}&.303&\textbf{.423}&.250\\
&\textit{VGG16}&.248&\textbf{.384}&.244&\textbf{.303}&.221&\textbf{.339}&.155\\
&\textit{VGG19}&.265&\textbf{.388}&.257&\textbf{.313}&.237&\textbf{.348}&.167\\
&\textit{Xception}&.387&\textbf{.564}&.428&\textbf{.551}&.347&\textbf{.524}&.251\\
\hline
\end{tabular}}
\caption{Insertion score for different models with explanation methods. We report the area under the curves formed by the originally predicted probability of modified images and the normalized probability (probability ratio: the predicted probability of modified images over the predicted probability of original images.) IDGI improves all methods for all models.}
\label{exp3}
\vspace{-1mm}
\end{table}

\subsection{Insertion Score}

\label{exp_insertion}
In this investigation, we assess our explanation approaches with the Insertion Score from prior works \cite{pan2021explaining, petsiuk2018rise}. For each test image, starting with a blank image, the insertion technique inserts pixels from the best to the lowest attribution scores and generates predictions. The approach generates a curve representing the prediction values as a function of the number of inserted pixels. We also compute the normalized curve where the curves are divided by the predicted class probability of the original image. The area under the curves (AUC) are then defined as the insertion scores. The quality of interpretation improves as the insertion score increases. For each image, we iteratively insert the next top 5\% important pixel values to the black base image based on each explanation technique, and we create the model performance curves as the mean model performance on all the reconstructed images at each level. We report the insertion scores in \cref{exp3}. As expected, the VG always has the lowest score across the experiments with different models. IDGI improves the Insertion Score drastically, in all cases, for all the IG-based methods.

\subsection{SIC and AIC}
\label{exp_sicaic}
We next evaluate the explanation methods using the Softmax information curves (SIC) and the Accuracy information curves (AIC) \cite{kapishnikov2019xrai}. The evaluation method starts with a blurred version of the given test image and then puts back the most important pixel's values as determined by the explanation approach, resulting in a bokeh image. Moreover, an information level is computed for each bokeh image by comparing the size of the compressed bokeh image to the size of the compressed original image, which is also referred to as the Normalized Entropy. Bokeh images are binned based on the information level. Then, the average accuracy for each bin is calculated. AIC is the curve of those mean accuracies over bins. Further, the probability of bokeh versus the original image is calculated for each image in each bin. SIC is the curve of the median of those values. Areas under the AIC and SIC curves are computed; better explanation methods should have higher values.

\noindent {\bf Implementation Details.} For each image, we first create a base image using PIL\footnote{https://pillow.readthedocs.io/en/stable/index.html} with Gaussian blur with a radius of 20 pixels. Then we use 25 percentile values, distributed from 0 to 100, as thresholds ($k$) to determine the important pixels at each threshold level. That is, given a test instance, we construct 25 bokeh images where each of them contains the top $k$ percent important pixel's original value with the rest pixels' value from the base image. Furthermore, we utilize the implementation from \cite{kapishnikov2019xrai} to compute the Normalized Entropy and record the model performance on each bokeh image. For information level bin size, we use 100. We report, in \cref{exp1}, the AUC under AIC and SIC curves across all baselines and our methods. Our method improves all three IG-based methods across all models drastically.
\begin{table}[!t]
\centering
\resizebox{\columnwidth}{!}{
\begin{tabular}{|c|c||c|c||c|c||c|c||c|}
\hline
\textbf{Metrics}&\textbf{Models}&\multicolumn{6}{|c|}{\textbf{IG-based Methods}}&{\textbf{Other}}\\
\cline{3-9}
&&IG&+Ours&GIG&+Ours&BlurIG&+Ours&VG\\
\hline

\cline{3-9}
\hline
\cline{3-9}
\hline
\cline{3-9}
\hline
\cline{3-9}
\hline
\multirow{8}{*}{\rotatebox[origin=c]{0}{\shortstack{AUC \\ AIC \\($\uparrow$)}}}&\textit{DenseNet121}&.161&\textbf{.300}&.141&\textbf{.252}&.192&\textbf{.230}&.087\\
&\textit{DenseNet169}&.160&\textbf{.288}&.154&\textbf{.254}&.181&\textbf{.216}&.089\\
&\textit{DenseNet201}&.185&\textbf{.307}&.182&\textbf{.269}&.213&\textbf{.246}&.110\\
&\textit{InceptionV3}&.203&\textbf{.343}&.189&\textbf{.338}&.266&\textbf{.301}&.127\\
&\textit{MobileNetV2}&.098&\textbf{.233}&.114&\textbf{.204}&.145&\textbf{.197}&.068\\
&\textit{ResNet50V2}&.162&\textbf{.253}&.162&\textbf{.248}&.189&\textbf{.210}&.108\\
&\textit{ResNet101V2}&.177&\textbf{.268}&.163&\textbf{.253}&.198&\textbf{.215}&.116\\
&\textit{ResNet151V2}&.186&\textbf{.281}&.165&\textbf{.258}&.205&\textbf{.229}&.112\\
&\textit{VGG16}&.145&\textbf{.244}&.141&\textbf{.199}&.181&\textbf{.222}&.108\\
&\textit{VGG19}&.153&\textbf{.263}&.150&\textbf{.219}&.204&\textbf{.240}&.117\\
&\textit{Xception}&.238&\textbf{.404}&.239&\textbf{.381}&.309&\textbf{.355}&.174\\

\hline
\hline

\multirow{8}{*}{\rotatebox[origin=c]{0}{\shortstack{AUC \\ SIC \\($\uparrow$)}}}&\textit{DenseNet121}&.054&\textbf{.228}&.036&\textbf{.157}&.085&\textbf{.134}&.015\\
&\textit{DenseNet169}&.052&\textbf{.230}&.045&\textbf{.170}&.083&\textbf{.130}&.016\\
&\textit{DenseNet201}&.068&\textbf{.241}&.058&\textbf{.183}&.109&\textbf{.155}&.019\\
&\textit{InceptionV3}&.087&\textbf{.294}&.061&\textbf{.286}&.171&\textbf{.232}&.029\\
&\textit{MobileNetV2}&.020&\textbf{.145}&.023&\textbf{.111}&.043&\textbf{.103}&.011\\
&\textit{ResNet50V2}&.077&\textbf{.210}&.067&\textbf{.201}&.099&\textbf{.158}&.025\\
&\textit{ResNet101V2}&.095&\textbf{.231}&.070&\textbf{.201}&.117&\textbf{.165}&.026\\
&\textit{ResNet151V2}&.101&\textbf{.249}&.065&\textbf{.212}&.122&\textbf{.177}&.025\\
&\textit{VGG16}&.046&\textbf{.166}&.039&\textbf{.104}&.082&\textbf{.141}&.021\\
&\textit{VGG19}&.046&\textbf{.177}&.041&\textbf{.115}&.098&\textbf{.151}&.023\\
&\textit{Xception}&.119&\textbf{.363}&.107&\textbf{.336}&.218&\textbf{.296}&.054\\

\hline
\end{tabular}}
\caption{Area under the curve for AIC and SIC with 11 different models. The information level is computed with the compression size ratio to the original image. IDGI shows improvement for all three IG-based methods across all experiment settings.}
\label{exp1}
\vspace{-1mm}
\end{table}

\begin{figure*}[!t]
     \centering
     \begin{subfigure}[b]{.88\textwidth}
         \centering
         \includegraphics[width=\textwidth]{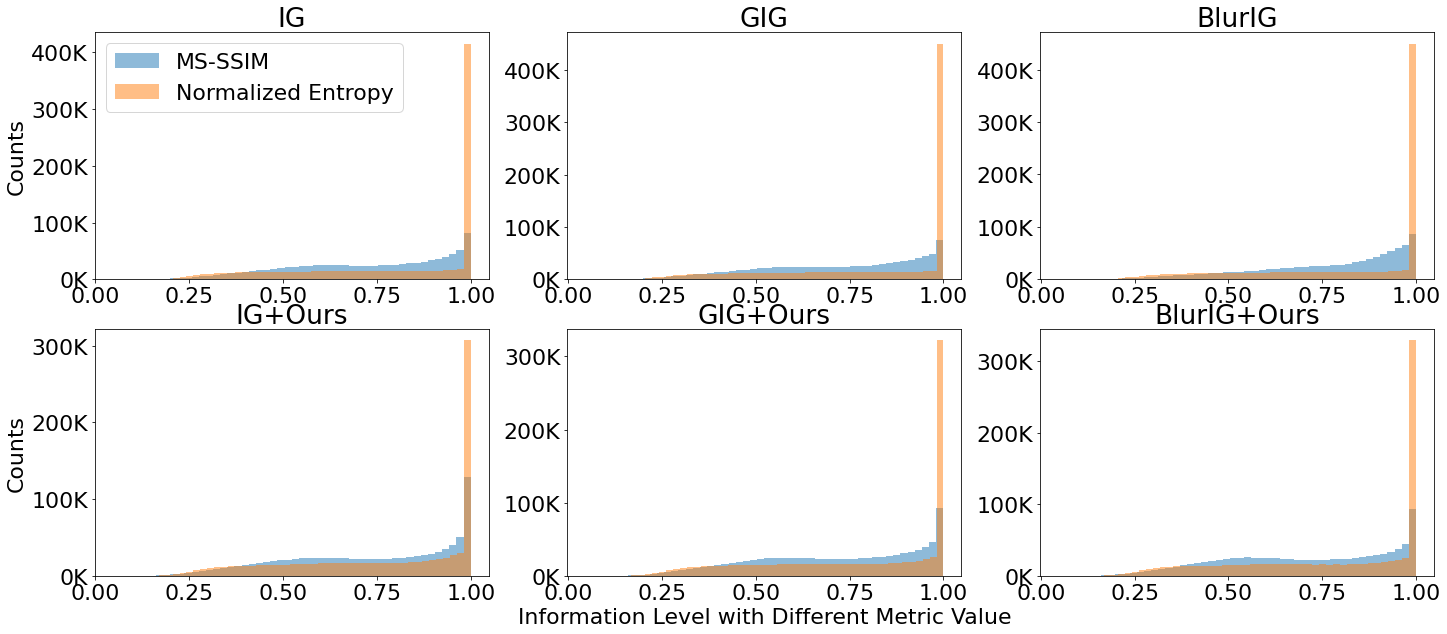}
     \end{subfigure}
     \vspace{-2mm}
\caption{Modified distribution of bokeh images over MS-SSIM and Normalized Entropy  \cite{kapishnikov2019xrai}. We produce 25 bokeh pictures for each test instance based on the various attribution maps and calculate the MS-SSIM and Normalized Entropy as the information level for each bokeh image. We show the modified bokeh images of the test instance for \textit{Resnet151V2}. Information level with MS-SSIM has a more evenly distribution of bokeh images than Normalized Entropy. Other models have similar trends.}
\label{fig_exp_dist}
\vspace{-4mm}
\end{figure*}

\begin{figure}[!t]
     \centering
     \vspace{-2mm}
     \begin{subfigure}[b]{.48\textwidth}
         \centering
         \includegraphics[width=\textwidth]{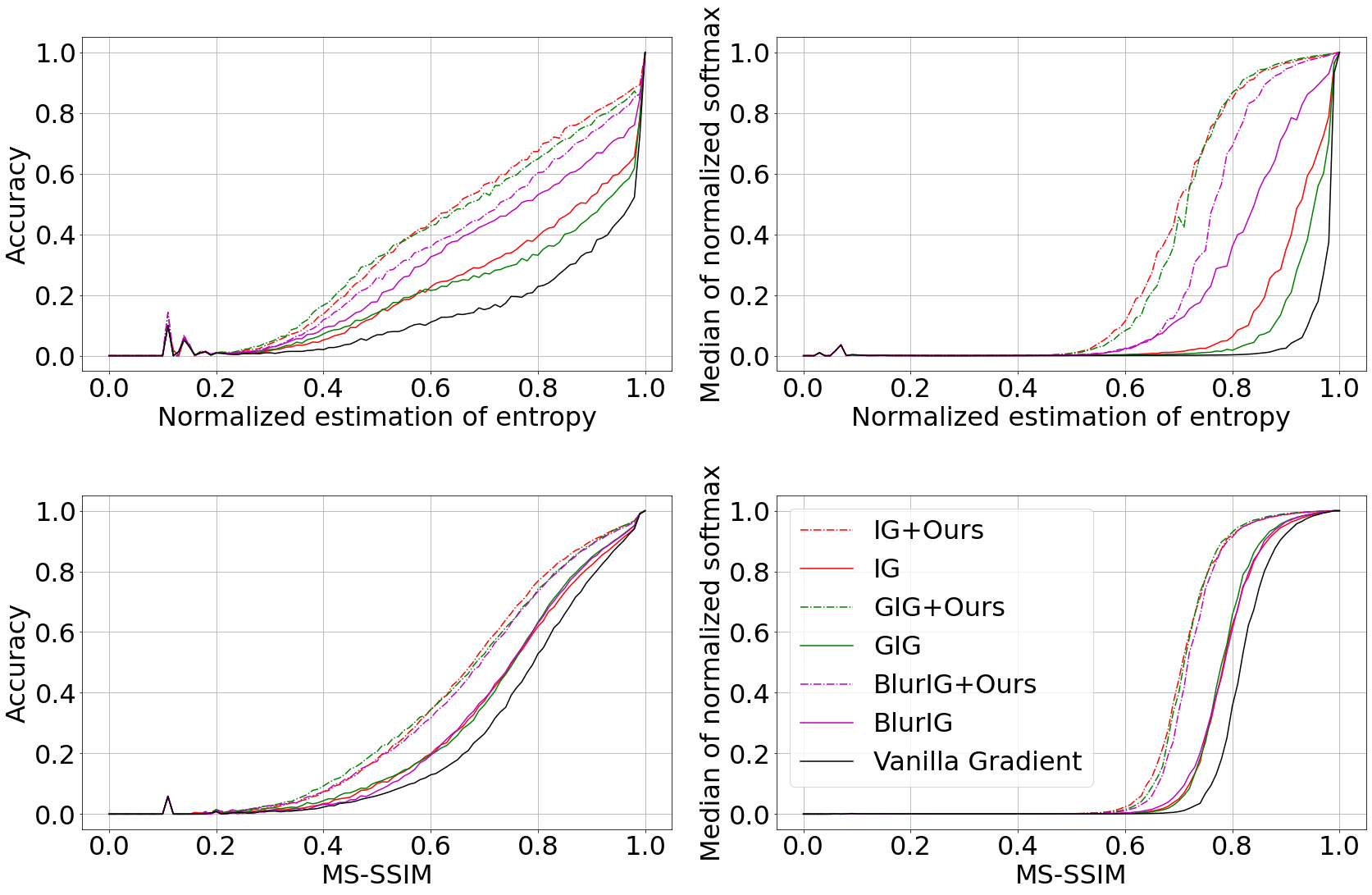}
     \end{subfigure}
     \vspace{-4mm}
\caption{AIC and SIC (median of normalized softmax), from \textit{InceptionV3} model, over the distribution of Normalized Entropy (Top row) and MS-SSIM (Bottom row). The area under the curve indicates IDGI provides significant improvement from all the IG-based methods. As expected, the saliency maps from the Vanilla gradient have the lowest score in all the experiments.}
\label{fig_exp2}
\end{figure}

\subsection{SIC and AIC with MS-SSIM}
\label{exp_sicaic_msssim}
In the third experiment, instead of the Normalized entropy as the information level, we use the Multi-scale Structural Similarity (MS-SSIM) \cite{wang2003multiscale}. MS-SSIM is a well-studied \cite{8451296, odena2017conditional, 7780553} image quality evaluation method that analyzes the structural similarity of two images. It is a multi-scale variation of a well-defined perceptual similarity measure that seeks to dismiss irrelevant features (based on human perspective) of a picture \cite{odena2017conditional}. MS-SSIM values range from 0.0 to 1.0; images with greater MS-SSIM values are perceptually more similar. \cref{fig_exp_dist} shows the distribution of all modified images over MS-SSIM and the Normalized Entropy, where MS-SSIM distributed the bokeh images more evenly over the bins. That is, each bin is expected to have more samples to represent the true model performance conditioned on that similarity. The effect can be also observed in \cref{fig_exp2} where the performance for the small Normalized Entropy group has high variance due to not having enough images for that group bin. 
In contrast, the measurement with MS-SSIM is a much smoother curve. Hence, we propose to use MS-SSIM for estimating the information of each bokeh image rather than Normalized Entropy.

The experimental settings are the same as in the previous section except we replaced the Normalized Entropy with the MS-SSIM score as the information level. As demonstrated in \cref{exp2}, our approach significantly improves modified AIC and SIC for all IG-based methods.

\begin{table}[!t]
\centering
\resizebox{\columnwidth}{!}{
\begin{tabular}{|c|c||c|c||c|c||c|c||c|}
\hline
\textbf{Metrics}&\textbf{Models}&\multicolumn{6}{|c|}{\textbf{IG-based Methods}}&{\textbf{Other}}\\
\cline{3-9}
&&IG&+Ours&GIG&+Ours&BlurIG&+Ours&VG\\
\hline

\cline{3-9}
\hline
\cline{3-9}
\hline
\cline{3-9}
\hline
\cline{3-9}
\hline
\multirow{8}{*}{\rotatebox[origin=c]{0}{\shortstack{AUC \\ AIC \\($\uparrow$)}}}&\textit{DenseNet121}&.229&\textbf{.305}&.231&\textbf{.280}&.216&\textbf{.277}&.186\\
&\textit{DenseNet169}&.241&\textbf{.314}&.249&\textbf{.297}&.218&\textbf{.289}&.205\\
&\textit{DenseNet201}&.254&\textbf{.323}&.262&\textbf{.303}&.237&\textbf{.303}&.216\\
&\textit{InceptionV3}&.264&\textbf{.333}&.268&\textbf{.333}&.264&\textbf{.323}&.228\\
&\textit{MobileNetV2}&.179&\textbf{.259}&.197&\textbf{.238}&.186&\textbf{.241}&.150\\
&\textit{ResNet50V2}&.225&\textbf{.277}&.239&\textbf{.274}&.209&\textbf{.260}&.198\\
&\textit{ResNet101V2}&.235&\textbf{.284}&.243&\textbf{.277}&.215&\textbf{.265}&.206\\
&\textit{ResNet151V2}&.247&\textbf{.302}&.250&\textbf{.292}&.227&\textbf{.284}&.212\\
&\textit{VGG16}&.205&\textbf{.271}&.212&\textbf{.245}&.204&\textbf{.259}&.179\\
&\textit{VGG19}&.211&\textbf{.275}&.220&\textbf{.252}&.214&\textbf{.266}&.188\\
&\textit{Xception}&.281&\textbf{.362}&.293&\textbf{.356}&.284&\textbf{.345}&.254\\

\hline
\hline

\multirow{8}{*}{\rotatebox[origin=c]{0}{\shortstack{AUC \\ SIC \\($\uparrow$)}}}&\textit{DenseNet121}&.184&\textbf{.263}&.188&\textbf{.239}&.172&\textbf{.236}&.139\\
&\textit{DenseNet169}&.205&\textbf{.282}&.214&\textbf{.263}&.182&\textbf{.256}&.166\\
&\textit{DenseNet201}&.212&\textbf{.286}&.221&\textbf{.265}&.194&\textbf{.266}&.170\\
&\textit{InceptionV3}&.211&\textbf{.287}&.215&\textbf{.285}&.214&\textbf{.276}&.179\\
&\textit{MobileNetV2}&.126&\textbf{.204}&.144&\textbf{.187}&.130&\textbf{.188}&.096\\
&\textit{ResNet50V2}&.196&\textbf{.254}&.213&\textbf{.250}&.177&\textbf{.236}&.167\\
&\textit{ResNet101V2}&.210&\textbf{.265}&.221&\textbf{.256}&.188&\textbf{.244}&.180\\
&\textit{ResNet151V2}&.221&\textbf{.282}&.227&\textbf{.270}&.197&\textbf{.261}&.186\\
&\textit{VGG16}&.163&\textbf{.234}&.174&\textbf{.210}&.166&\textbf{.224}&.137\\
&\textit{VGG19}&.173&\textbf{.240}&.186&\textbf{.219}&.177&\textbf{.233}&.149\\
&\textit{Xception}&.223&\textbf{.312}&.233&\textbf{.304}&.229&\textbf{.293}&.194\\
\hline
\end{tabular}}
\caption{Area under the curve for SIC and AIC for 11 models. The information level is MS-SSIM. IDGI shows improvement for all three
IG-based methods across all experiments.}
\label{exp2}

\end{table}

\subsection{Weakly Supervised Localization}
\label{exp_f1rocmae}
We utilize the quantitative assessments in \cite{cong2018review, xu2020attribution, kapishnikov2019xrai}. The evaluation computes ROC-AUC, F1, and MAE (mean absolute error) of the created saliency mask by considering pixels inside the annotation to be positive and pixels outside to be negative, given an annotation area. Specifically, we calculate the optimal F1 and MAE of each saliency map for each image by finding the best threshold for attribution values provided by different explanation methods. \cref{exp4} summarizes the findings for each metric where IDGI consistently outperforms the underlying IG-based approaches.

\begin{table}[!t]
\centering
\resizebox{\columnwidth}{!}{
\begin{tabular}{|c|c||c|c||c|c||c|c||c|}
\hline
\textbf{Metrics}&\textbf{Models}&\multicolumn{6}{|c|}{\textbf{IG-based Methods}}&{\textbf{Other}}\\
\cline{3-9}
&&IG&+Ours&GIG&+Ours&BlurIG&+Ours&VG\\
\hline

\cline{3-9}
\hline
\cline{3-9}
\hline
\cline{3-9}
\hline
\cline{3-9}
\hline
\multirow{8}{*}{\rotatebox[origin=c]{0}{\shortstack{F1 \\($\uparrow$)}}}&\textit{DenseNet121}&.663&\textbf{.733}&.676&\textbf{.700}&.664&\textbf{.694}&.648\\
&\textit{DenseNet169}&.666&\textbf{.730}&.679&\textbf{.702}&.667&\textbf{.695}&.649\\
&\textit{DenseNet201}&.658&\textbf{.723}&.675&\textbf{.701}&.661&\textbf{.694}&.645\\
&\textit{InceptionV3}&.661&\textbf{.731}&.679&\textbf{.727}&.670&\textbf{.717}&.651\\
&\textit{MobileNetV2}&.666&\textbf{.741}&.686&\textbf{.713}&.673&\textbf{.718}&.656\\
&\textit{ResNet50V2}&.673&\textbf{.724}&.694&\textbf{.720}&.676&\textbf{.704}&.668\\
&\textit{ResNet101V2}&.675&\textbf{.730}&.691&\textbf{.716}&.676&\textbf{.703}&.666\\
&\textit{ResNet151V2}&.675&\textbf{.730}&.687&\textbf{.713}&.675&\textbf{.701}&.663\\
&\textit{VGG16}&.672&\textbf{.719}&.671&\textbf{.696}&.672&\textbf{.704}&.667\\
&\textit{VGG19}&.672&\textbf{.719}&.671&\textbf{.696}&.671&\textbf{.702}&.666\\
&\textit{Xception}&.669&\textbf{.745}&.691&\textbf{.740}&.678&\textbf{.730}&.662\\
\cline{2-9}

\hline
\hline

\multirow{8}{*}{\rotatebox[origin=c]{0}{\shortstack{ROC \\ AUC \\($\uparrow$)}}}&\textit{DenseNet121}&.662&\textbf{.798}&.660&\textbf{.722}&.661&\textbf{.722}&.607\\
&\textit{DenseNet169}&.656&\textbf{.790}&.655&\textbf{.714}&.657&\textbf{.718}&.582\\
&\textit{DenseNet201}&.654&\textbf{.788}&.664&\textbf{.729}&.658&\textbf{.726}&.604\\
&\textit{InceptionV3}&.679&\textbf{.811}&.664&\textbf{.799}&.696&\textbf{.790}&.659\\
&\textit{MobileNetV2}&.677&\textbf{.811}&.695&\textbf{.760}&.684&\textbf{.778}&.653\\
&\textit{ResNet50V2}&.698&\textbf{.798}&.698&\textbf{.775}&.690&\textbf{.746}&.671\\
&\textit{ResNet101V2}&.709&\textbf{.811}&.693&\textbf{.773}&.695&\textbf{.753}&.670\\
&\textit{ResNet151V2}&.707&\textbf{.810}&.681&\textbf{.766}&.687&\textbf{.745}&.659\\
&\textit{VGG16}&.652&\textbf{.757}&.648&\textbf{.702}&.656&\textbf{.727}&.642\\
&\textit{VGG19}&.652&\textbf{.755}&.650&\textbf{.703}&.652&\textbf{.719}&.640\\
&\textit{Xception}&.693&\textbf{.825}&.702&\textbf{.814}&.705&\textbf{.805}&.683\\
\cline{2-9}

\hline
\hline

\multirow{8}{*}{\rotatebox[origin=c]{0}{\shortstack{MAE \\($\downarrow$)}}}&\textit{DenseNet121}&.236&\textbf{.189}&.232&\textbf{.215}&.238&\textbf{.218}&.251\\
&\textit{DenseNet169}&.239&\textbf{.195}&.235&\textbf{.218}&.241&\textbf{.222}&.257\\
&\textit{DenseNet201}&.237&\textbf{.194}&.230&\textbf{.212}&.238&\textbf{.216}&.251\\
&\textit{InceptionV3}&.233&\textbf{.187}&.228&\textbf{.188}&.227&\textbf{.194}&.241\\
&\textit{MobileNetV2}&.234&\textbf{.183}&.225&\textbf{.203}&.231&\textbf{.199}&.242\\
&\textit{ResNet50V2}&.233&\textbf{.195}&.223&\textbf{.200}&.234&\textbf{.213}&.240\\
&\textit{ResNet101V2}&.229&\textbf{.189}&.223&\textbf{.201}&.232&\textbf{.212}&.240\\
&\textit{ResNet151V2}&.230&\textbf{.189}&.227&\textbf{.204}&.234&\textbf{.214}&.243\\
&\textit{VGG16}&.239&\textbf{.206}&.241&\textbf{.225}&.240&\textbf{.217}&.244\\
&\textit{VGG19}&.239&\textbf{.206}&.241&\textbf{.224}&.241&\textbf{.219}&.245\\
&\textit{Xception}&.227&\textbf{.178}&.218&\textbf{.179}&.222&\textbf{.186}&.233\\
\cline{2-9}

\hline
\end{tabular}}
\caption{F1, ROC-AUC, and MAE scores.  IG-based methods are improved by IDGI across all three metrics and eleven models.}
\label{exp4}
\end{table}

\section{Related Work 
}
\label{sec_rel}
Research on explanations for machine learning models has garnered a significant amount of attention since the development of expert models of the 1970s \cite{ghorbani2019interpretation, adebayo2018sanity}. Recently, with the increased use of Deep Neural Networks, several papers have focused on explaining the decisions of these black box models. One possible approach is based on Shapley values \cite{shapley1953quota, gale1962college}. The Shapley value was proposed to represent the contribution value of each player in the cooperative games to the outcome of the game. From the explanation perspective, the Shapley value based methods \cite{vstrumbelj2014explaining} computed for each feature how much it contributes to the prediction score when it's considered alone compared to the rest of the features. The Shapley value of a feature is its true attribution value. However, calculating Shapley values is intractable when the input dimension is large. Several methods approximate the Shapley values. These include KernelSHAP \cite{lundberg2017unified}, BShap \cite{sundararajan2020many}, and FastShap \cite{jethani2021fastshap}.

Another strategy for explaining models' decisions is input perturbation. Input perturbation methods work by manipulating the input and observing its effect on the output; this process is often repeated many times to produce the general behavior of the model's prediction on that input. For example, LIME \cite{ribeiro2016should} approximates the decision for an input by fitting a linear model around the neighborhood of the input, where the neighborhood is generated through perturbations. The similar idea of manipulating the input is utilized by RISE \cite{petsiuk2018rise} and several other papers\cite{dabkowski2017real, fong2017interpretable, zintgraf2017visualizing}.

For deep neural networks, in addition to the above strategies, several have also focused on methods of backpropagation to assign attribution to the input values. For example, modified backpropagation based methods propagate the signal of the final prediction back to the input and assign a score for each feature in the input. Methods include Deconvnet \cite{zeiler2014visualizing}, guided backpropagation \cite{springenberg2014striving}, DeepLIFT \cite{shrikumar2017learning} and LRP \cite{bach2015pixel, montavon2019layer}. These approaches propagate the modified gradient/signal, instead of the true gradient, to the input. 

Using the true gradient of the prediction with respect to the input as the explanation was introduced by Simonyan et al.\cite{simonyan2013deep}. Similarly, Shrikumar et al.\cite{shrikumar2017learning} propose using the element-wise product (gradient $\otimes$ input) with input itself instead of the gradient directly. Grad-CAM\cite{selvaraju2017grad} utilizes gradients to produce the localization map of the important regions in the input image. A popular method, Integrated Gradients (IG), was proposed by Sundararajan et al.\cite{sundararajan2017axiomatic}; it computes the attribution for each feature to explain the decision of a differentiable model, e.g. DNNs. Its variants include Guided Integrated Gradients (GIG)\cite{kapishnikov2021guided} and Blur Integrated Gradients (BlurIG)\cite{xu2020attribution}. XRAI \cite{kapishnikov2019xrai} utilizes IG to provide interpretation at the region level instead of the pixel level. SmoothGrad \cite{smilkov2017smoothgrad} and AGI \cite{pan2021explaining} compute repeated attribution maps for a given input; however, those approaches require more computation as the single path methods (such as IG, GIG, and BlurIG) has to be repeated multiple times for a given input. 
Finally, I-GOS \cite{qi2019visualizing} and I-GOS++\cite{khorram2021igos++} find a mask that would decrease the prediction score the most when it is applied to an image; they use the output of IG in their search for the mask. Our work builds on, and is orthogonal to, the single-path IG-based methods where we reduce the noise created during the integration computation. In other words, our work can be applied to any given single-path IG-based method. Furthermore, since it works with single-path methods, it is also easy to adapt to multiple-path methods as well.

%%%%%%%%% Conclusion
\section{Conclusion}
\label{sec_con}

We investigate the noise source generated by Integrated Gradient (IG) and its variants. Specifically, we propose the Important Direction Gradient Integration (IDGI) framework, which can be incorporated into all IG-based explanation methods and reduce the noise in their outputs. Extensive experiments show that IDGI can drastically improve the quality of saliency maps generated by the underlying IG-based approaches.

\noindent {\bf Acknowledgements.} {This work of Wang was supported by Wang's startup funding, the Cisco Research Award, and the National Science
Foundation under grant No.~2216926.}

\newpage
%%%%%%%%% REFERENCES
{\small
\bibliographystyle{ieee_fullname}
\bibliography{egbib}
}

\end{document}

\title{IDGI: A Framework to Eliminate  Explanation Noise from Gradients Integration (Appendix)}

\maketitle
\section{Theorem}
\begin{theorem}
\label{theorem_1}
    Given a function $f_c(x): R^n \rightarrow R$, points $x_j, x_{j+1}, x_{j_p} \in R^n$, then the gradient of the function with respect to each point in the space $R^n$ forms the conservative vector fields $\overrightarrow{F}$ and further define the hyperplane $h_j=\{x: f_c(x)=f_c(x_j)\}$ in $\overrightarrow{F}$. Assume the Riemann Integration accurately estimates the line integral of the vector field $\overrightarrow{F}$ from points $x_j$ to $x_{j+1}$ and $x_{j_p}$ e.g. $\int_{x_j}^{x_{j_p}}\frac{\partial f_c(x)}{\partial x} dx \approx \frac{\partial f_c(x_j)}{\partial x_j}{(x_{j_p} - x_j)}$, and $x_j \in h_j$, $x_{j_p}, x_{j+1} \in h_{j+1}$. Then:
    \[\int_{x_j}^{x_{j+1}}\frac{\partial f_c(x)}{\partial x} dx \approx \int_{x_j}^{x_{j_p}}\frac{\partial f_c(x)}{\partial x} dx.\]
\end{theorem}
\textit{Proof.}
\begin{align}
    \int_{x_j}^{x_{j+1}}\frac{\partial f_c(x)}{\partial x} dx &= f_c(x_{j+1}) - f_c(x_{j}) \nonumber \\
    & \approx \frac{\partial f_c(x_j)}{\partial x_j}{(x_{j+1} - x_j)} \nonumber \\
    &= f_c(x_{j_p}) - f_c(x_{j}) \nonumber \\
    &=\frac{\partial f_c(x_j)}{\partial x_j}{(x_{j_p} - x_j)} \nonumber \\
    &=\int_{x_j}^{x_{j_p}}\frac{\partial f_c(x)}{\partial x} dx
\end{align}

\section{IG with IDGI}

\begin{figure}[h]
     \centering
     \begin{subfigure}[h]{0.15\textwidth}
         \includegraphics[width=\textwidth]{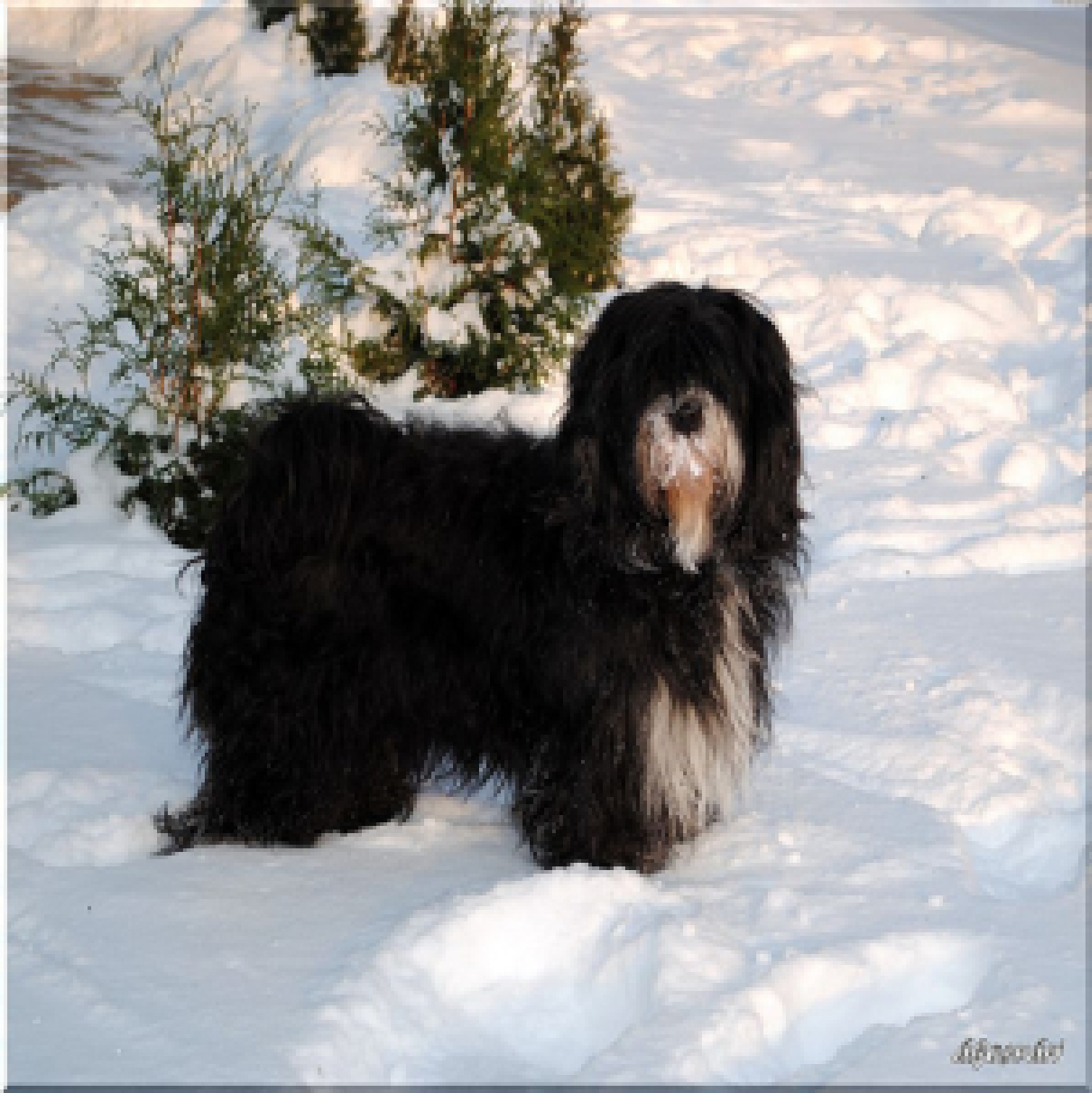}
         \caption{Original Image}
         \label{original_black}
     \end{subfigure}
     \begin{subfigure}[h]{0.15\textwidth}
         \includegraphics[width=\textwidth]{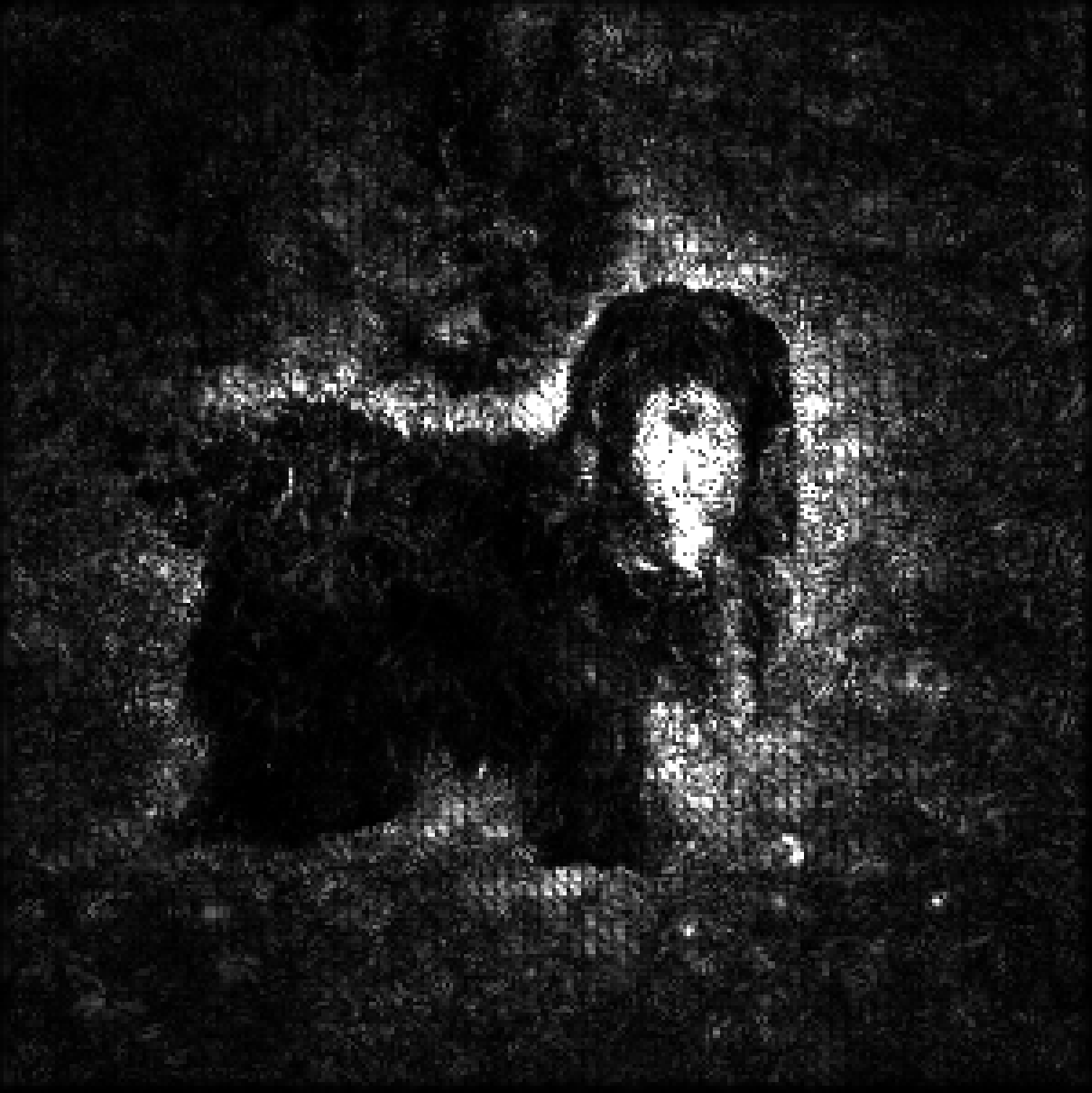}
         \caption{IG}
         \label{id_black}
     \end{subfigure}
     \begin{subfigure}[h]{.15\textwidth}
         \includegraphics[width=\textwidth]{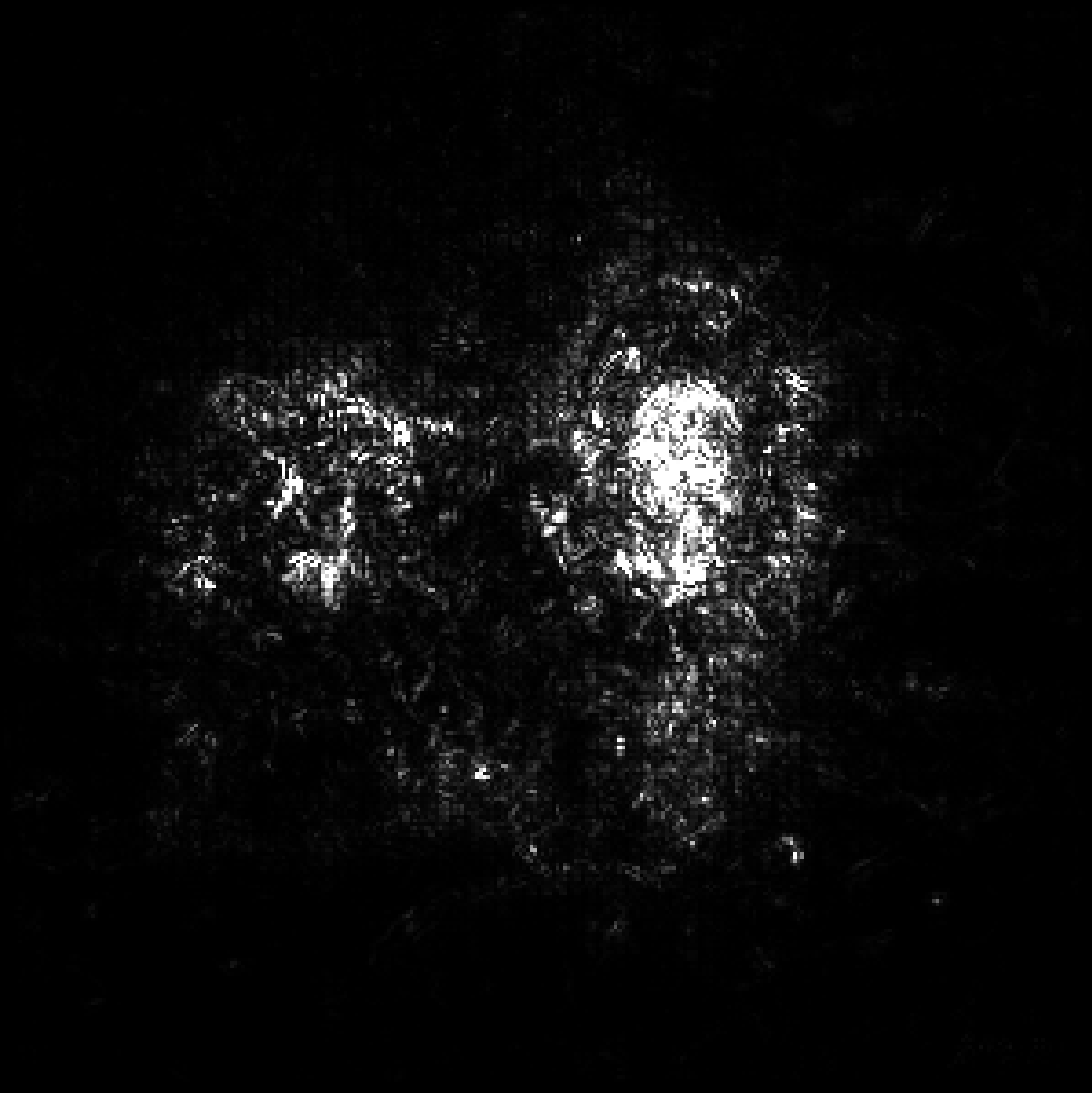}
         \caption{IG+IDGI}
         \label{id_idgi_black}
     \end{subfigure}
\caption{Original image is predicted \textit{Tibetan terrier} from Xception classifier. Both \ref{id_black} and \ref{id_idgi_black} are attributions from IG and IG+IDGI with the black image as reference. Since the pixels are also black for the original image on the dog region, by design, IG is not able to assign important values to those pixels, however, ID+IDGI overcomes the issue.}
\label{fig_black_image}
\end{figure}

The Integrated Gradients algorithm requires a specified reference image to compute the attribution. One often selects a black or white picture as a reference point, resulting in the zero attribution value to pixels with the same value as the reference. This is due to the fact that these pixel values do not change while traveling from the reference image to the original image. However, these pixels may still be crucial for the classifier to make the decision, and merit attribution differs from zero. For example, as shown in Figure \ref{fig_black_image}, the \textit{Xception} model makes the prediction correctly on the given image has a black dog. When utilizing IG for providing an explanation, the body of the dog will be assigned zero attributions since the reference (black) image has the same pixel value as the dog. Intuitively, the explanation method should give non-zero values to these black pixels, since they represent the dog's body and are assumed to be significant characteristics. Alternatively, if the attribution value is zero, it is likely because the feature is insignificant and not because of the explanation method's design. In contrast to the original IG, IG with IDGI might potentially assign non-zero values to pixels with the same value as the reference picture, a desirable trait for a superior explanation technique.

\section{Visual Examples}
\begin{figure*}[h]
     \centering
     \begin{subfigure}[b]{\textwidth}
         \centering
         \includegraphics[width=\textwidth]{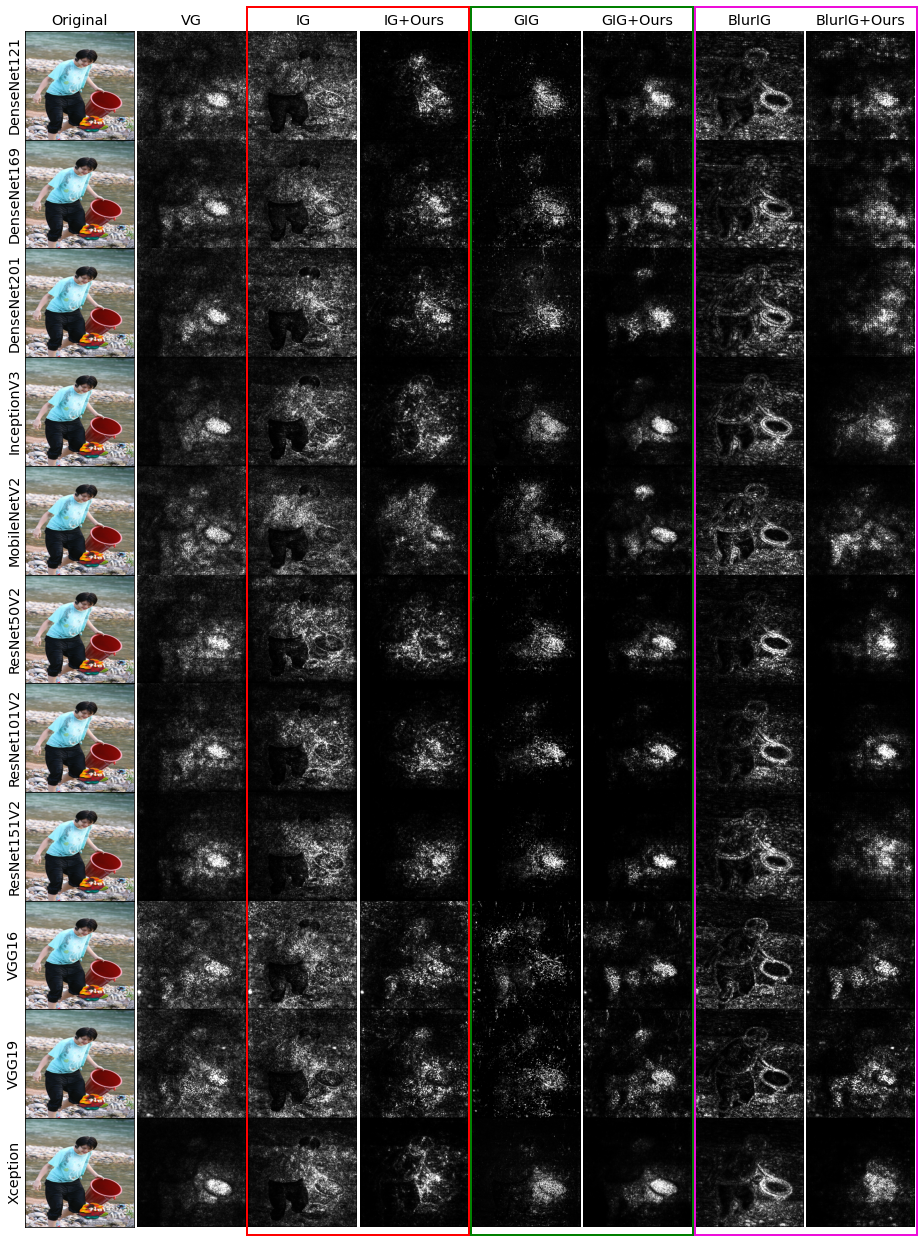}
     \end{subfigure}
\caption{Predicted Label for all models: \textit{bucket}
}
\label{q1}
\end{figure*}

\begin{figure*}[h]
     \centering
     \begin{subfigure}[b]{\textwidth}
         \centering
         \includegraphics[width=\textwidth]{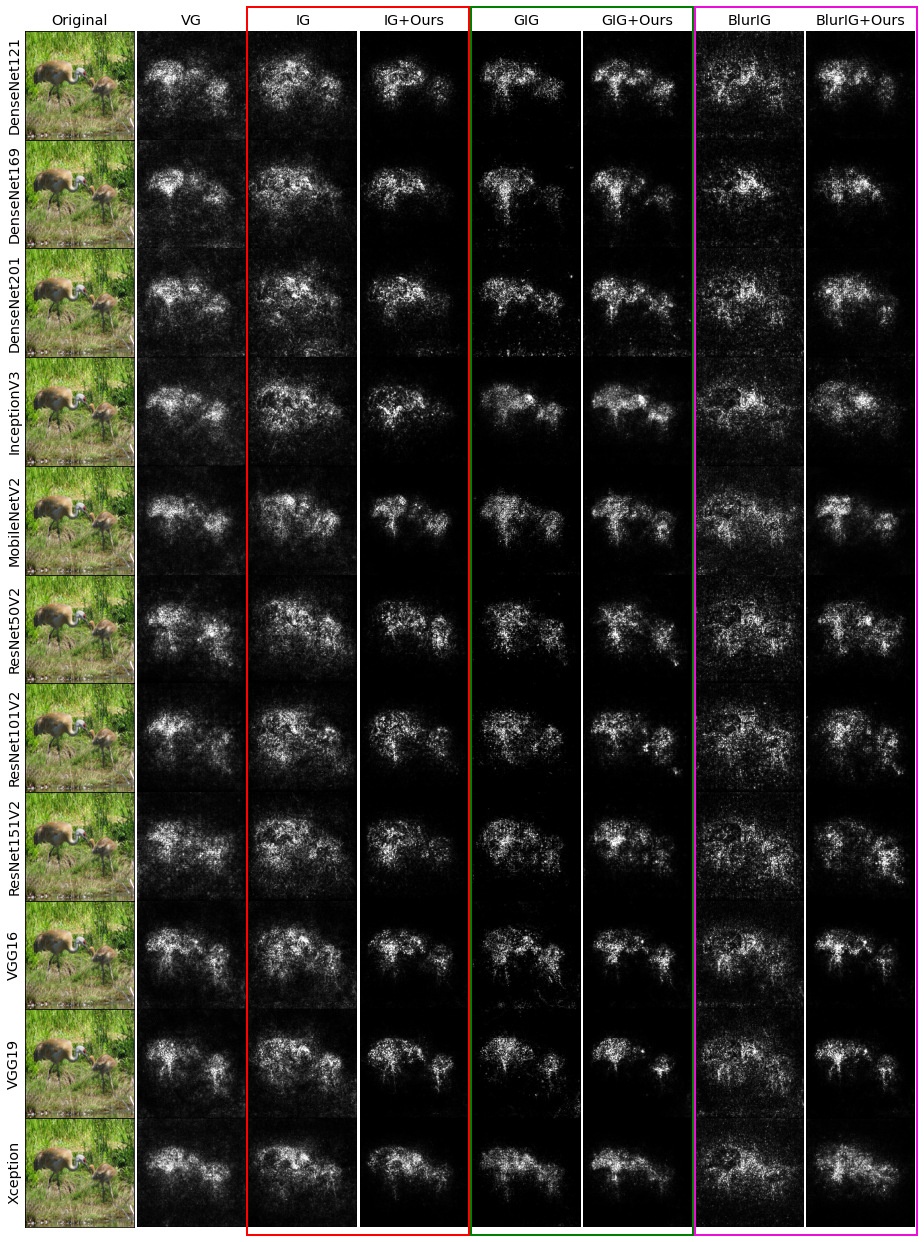}
     \end{subfigure}
\caption{Predicted Label for all models: \textit{crane}
}
\label{q2}
\end{figure*}

\begin{figure*}[h]
     \centering
     \begin{subfigure}[b]{\textwidth}
         \centering
         \includegraphics[width=\textwidth]{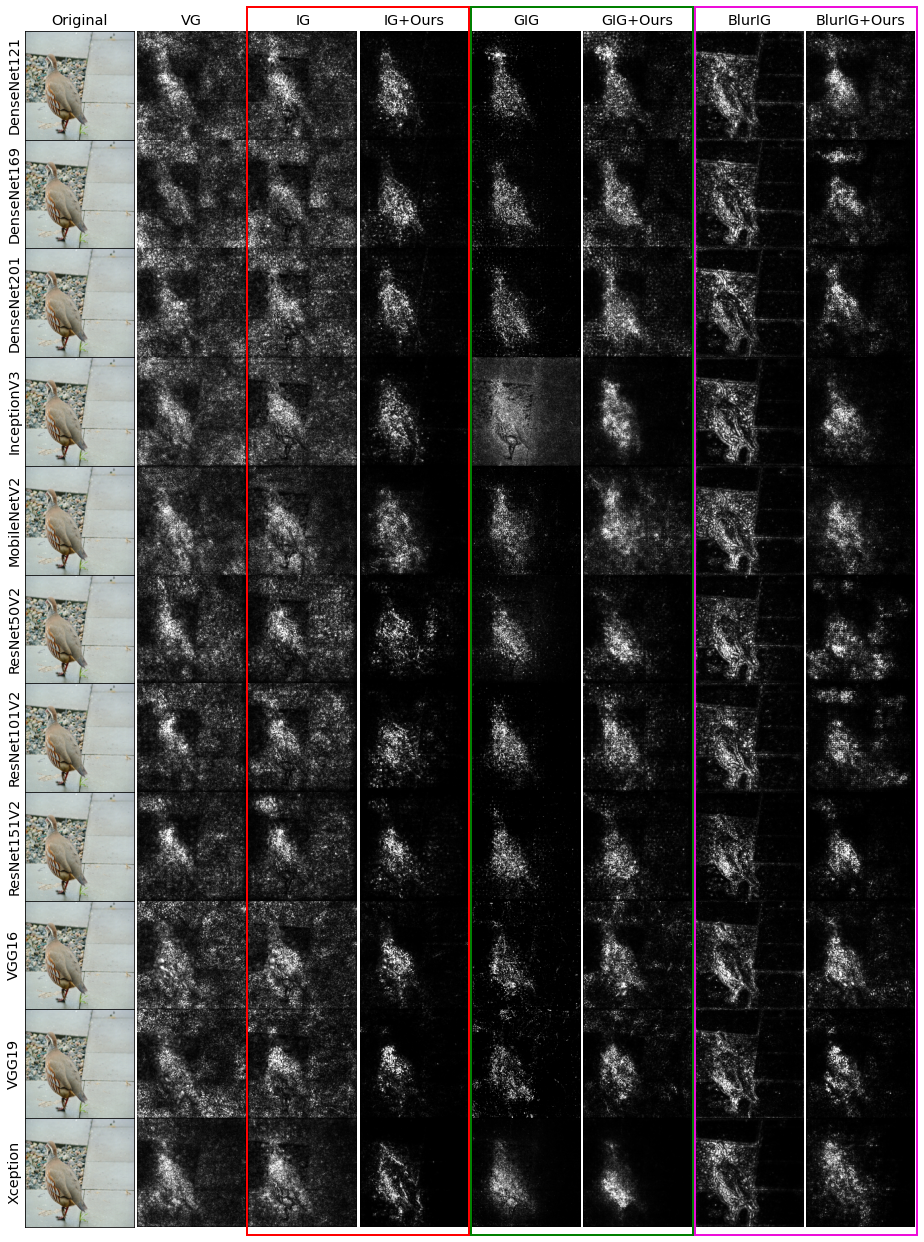}
     \end{subfigure}
\caption{Predicted Label for all models: \textit{partridge}
}
\label{q4}
\end{figure*}

\begin{figure*}[h]
     \centering
     \begin{subfigure}[b]{\textwidth}
         \centering
         \includegraphics[width=\textwidth]{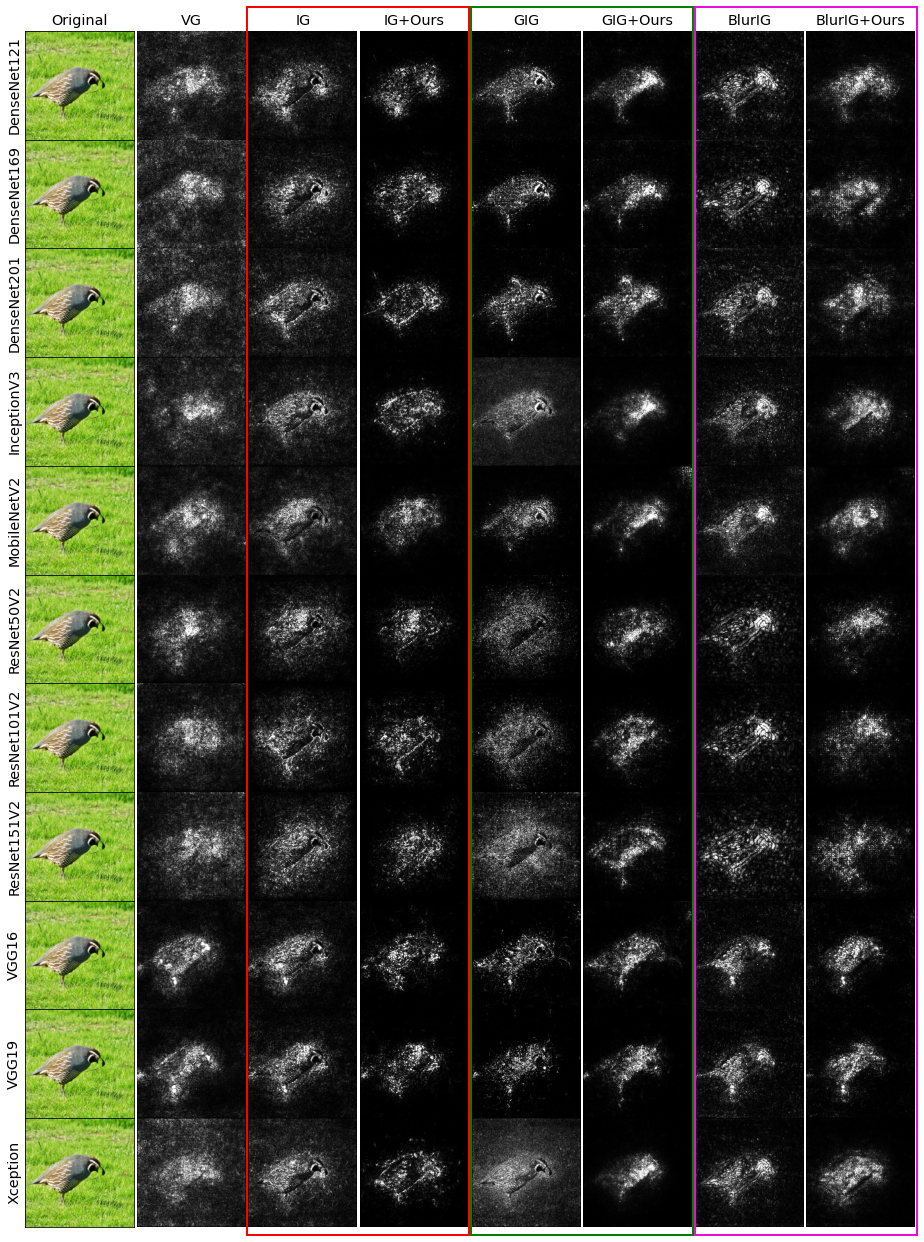}
     \end{subfigure}
\caption{Predicted Label for all models: \textit{quail}
}
\label{q5}
\end{figure*}

We present more visual examples in \cref{q1,q2,q3,q4,q5}.

\begin{figure*}[h]
     \centering
     \begin{subfigure}[b]{\textwidth}
         \centering
         \includegraphics[width=\textwidth]{figs/appendix_examples/mergus serrator.png}
     \end{subfigure}
\caption{Predicted Label for all models: \textit{mergus serrator}
}
\label{q3}
\end{figure*}

\section{Distribution by Normalized Entropy and MS-SSIM}
\begin{figure*}[!t]
     \centering
     \begin{subfigure}[b]{.9\textwidth}
         \centering
         \includegraphics[width=\textwidth]{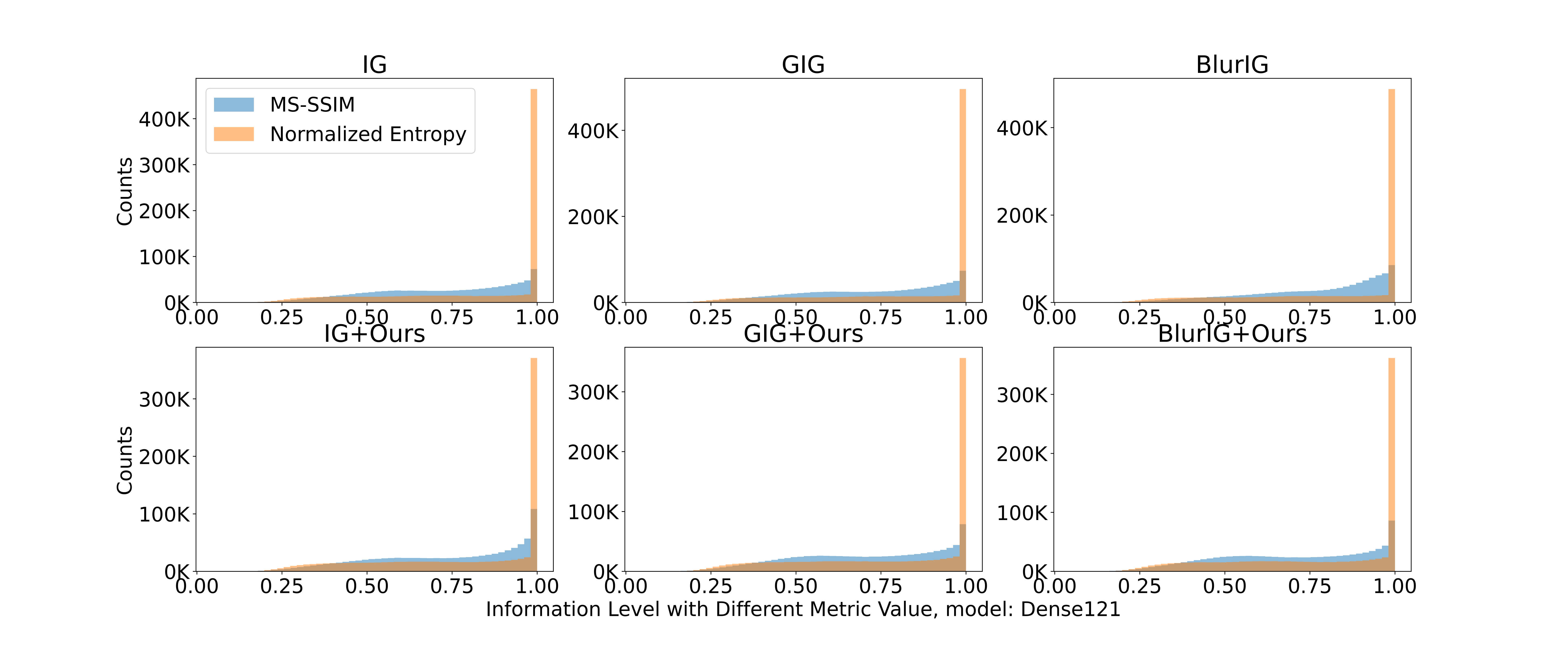}
     \end{subfigure}
\caption{Modified distribution of bokeh images over MS-SSIM and Normalized Entropy  \cite{kapishnikov2019xrai}. Model: \textit{DenseNet121} 
}
\label{Dense121_f}
\end{figure*}

\begin{figure*}[!t]
     \centering
     \begin{subfigure}[b]{.9\textwidth}
         \centering
         \includegraphics[width=\textwidth]{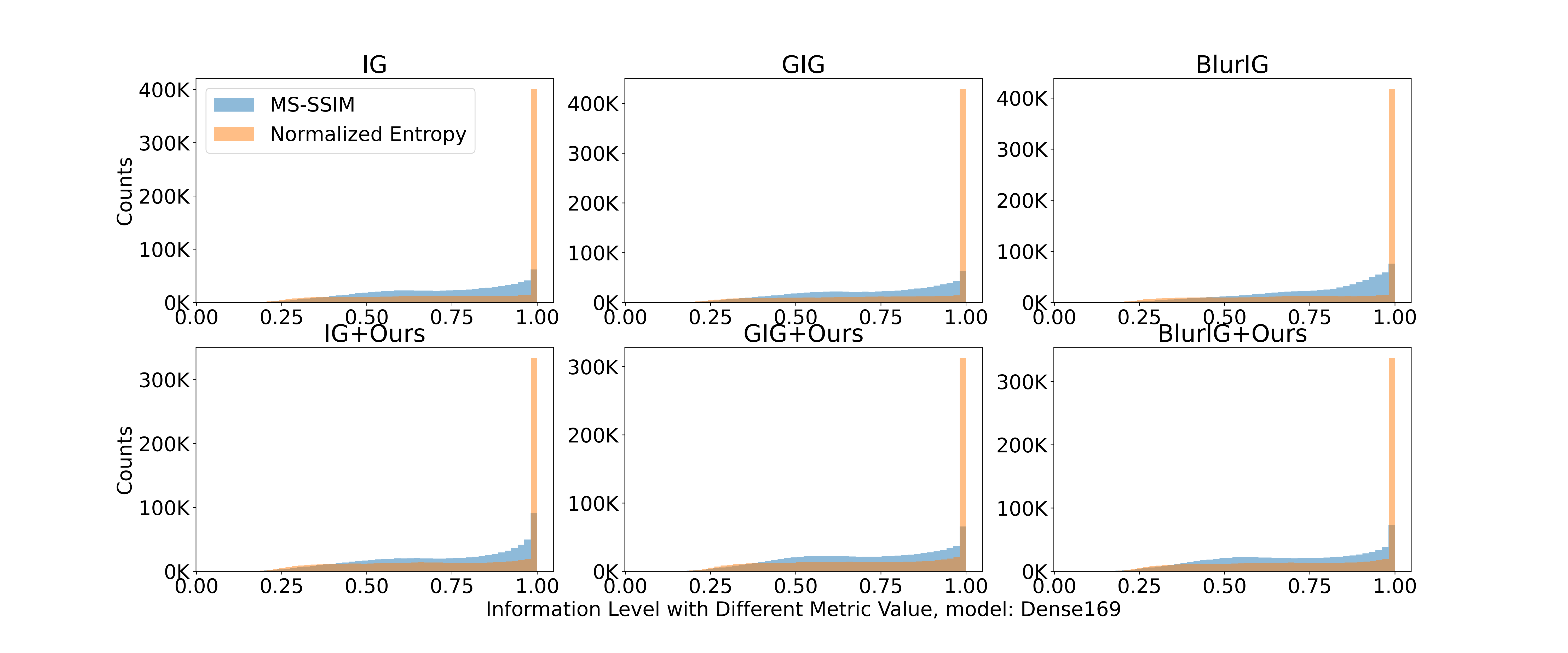}
     \end{subfigure}
\caption{Modified distribution of bokeh images over MS-SSIM and Normalized Entropy  \cite{kapishnikov2019xrai}. Model: \textit{DenseNet169} 
}
\label{Dense169_f}
\end{figure*}

\begin{figure*}[!t]
     \centering
     \begin{subfigure}[b]{.9\textwidth}
         \centering
         \includegraphics[width=\textwidth]{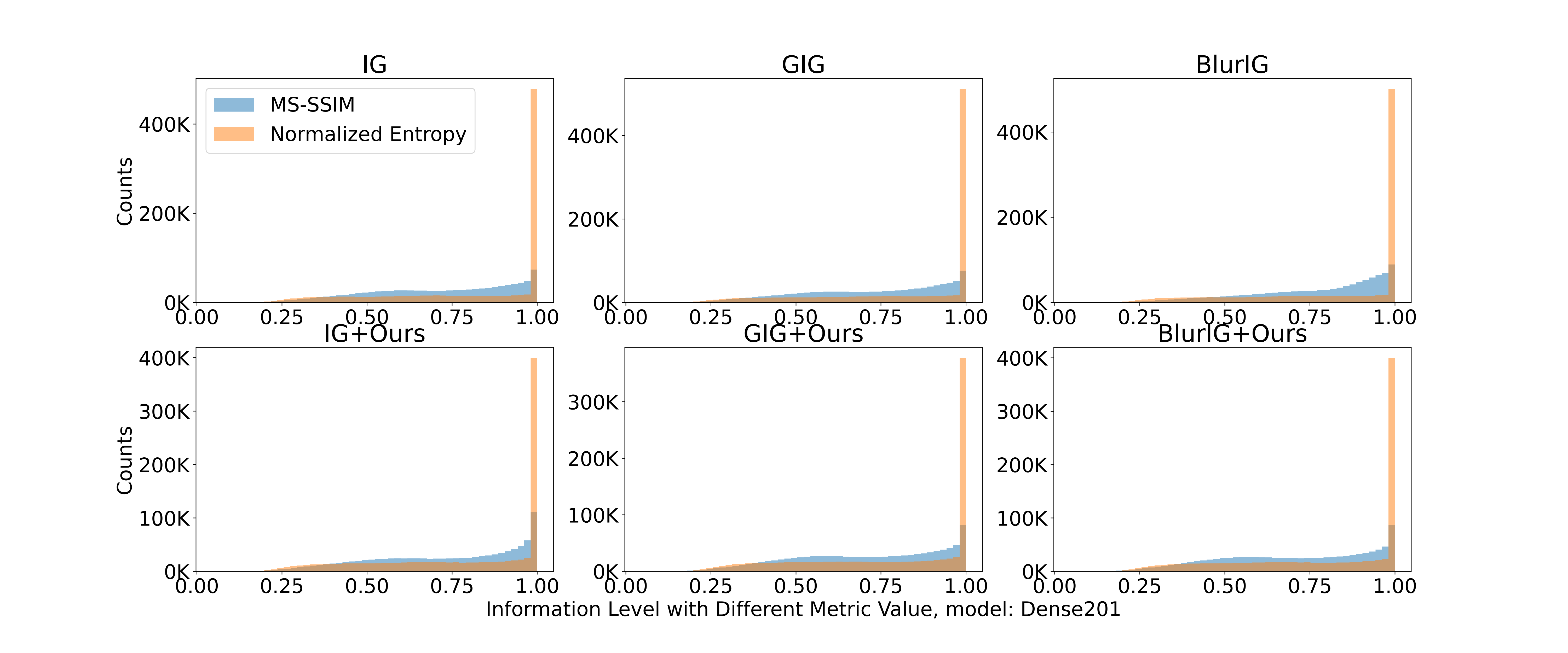}
     \end{subfigure}
\caption{Modified distribution of bokeh images over MS-SSIM and Normalized Entropy  \cite{kapishnikov2019xrai}. Model: \textit{DenseNet201} 
}
\label{Dense201_f}
\end{figure*}

\begin{figure*}[!t]
     \centering
     \begin{subfigure}[b]{.9\textwidth}
         \centering
         \includegraphics[width=\textwidth]{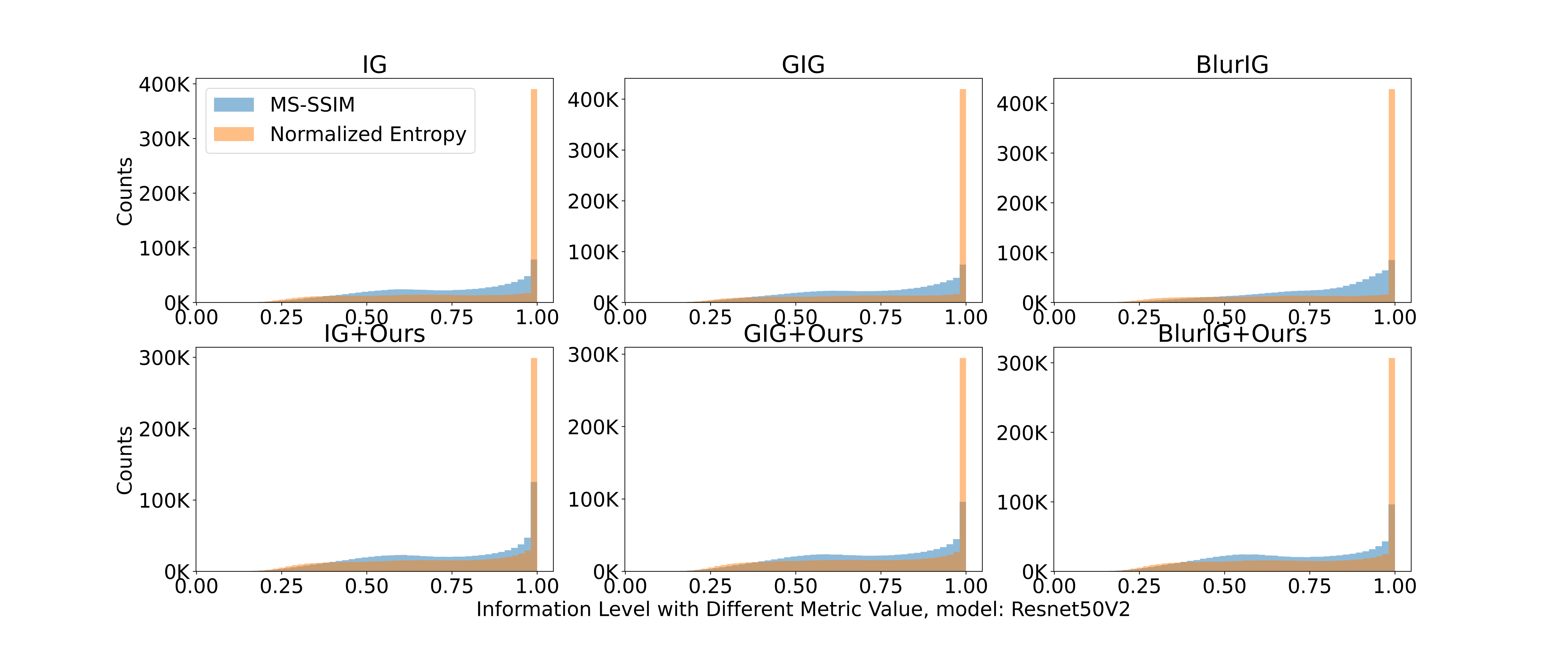}
     \end{subfigure}
\caption{Modified distribution of bokeh images over MS-SSIM and Normalized Entropy  \cite{kapishnikov2019xrai}. Model: \textit{Resnet50V2} 
}
\label{Resnet50V2_f}
\end{figure*}

\begin{figure*}[!t]
     \centering
     \begin{subfigure}[b]{.9\textwidth}
         \centering
         \includegraphics[width=\textwidth]{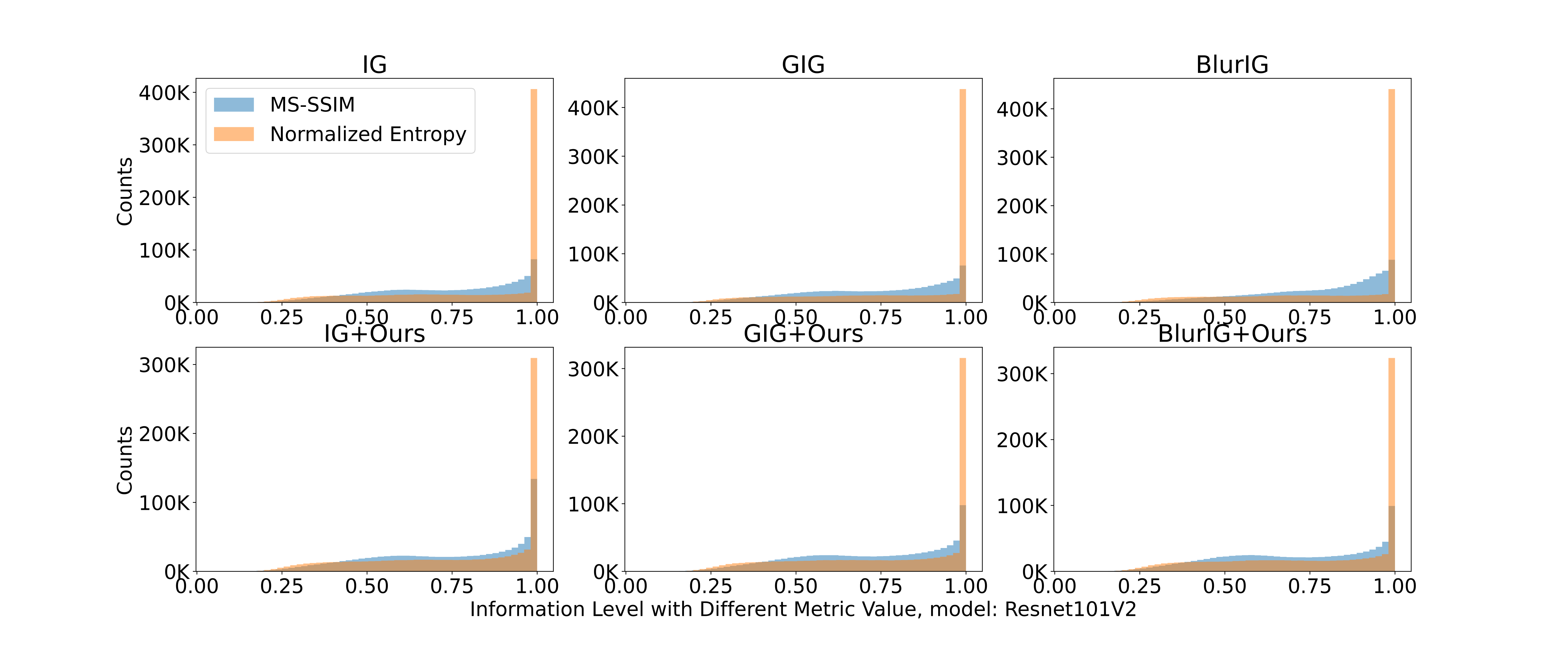}
     \end{subfigure}
\caption{Modified distribution of bokeh images over MS-SSIM and Normalized Entropy  \cite{kapishnikov2019xrai}. Model: \textit{Resnet101V2} 
}
\label{Resnet101V2_f}
\end{figure*}

\begin{figure*}[!t]
     \centering
     \begin{subfigure}[b]{.9\textwidth}
         \centering
         \includegraphics[width=\textwidth]{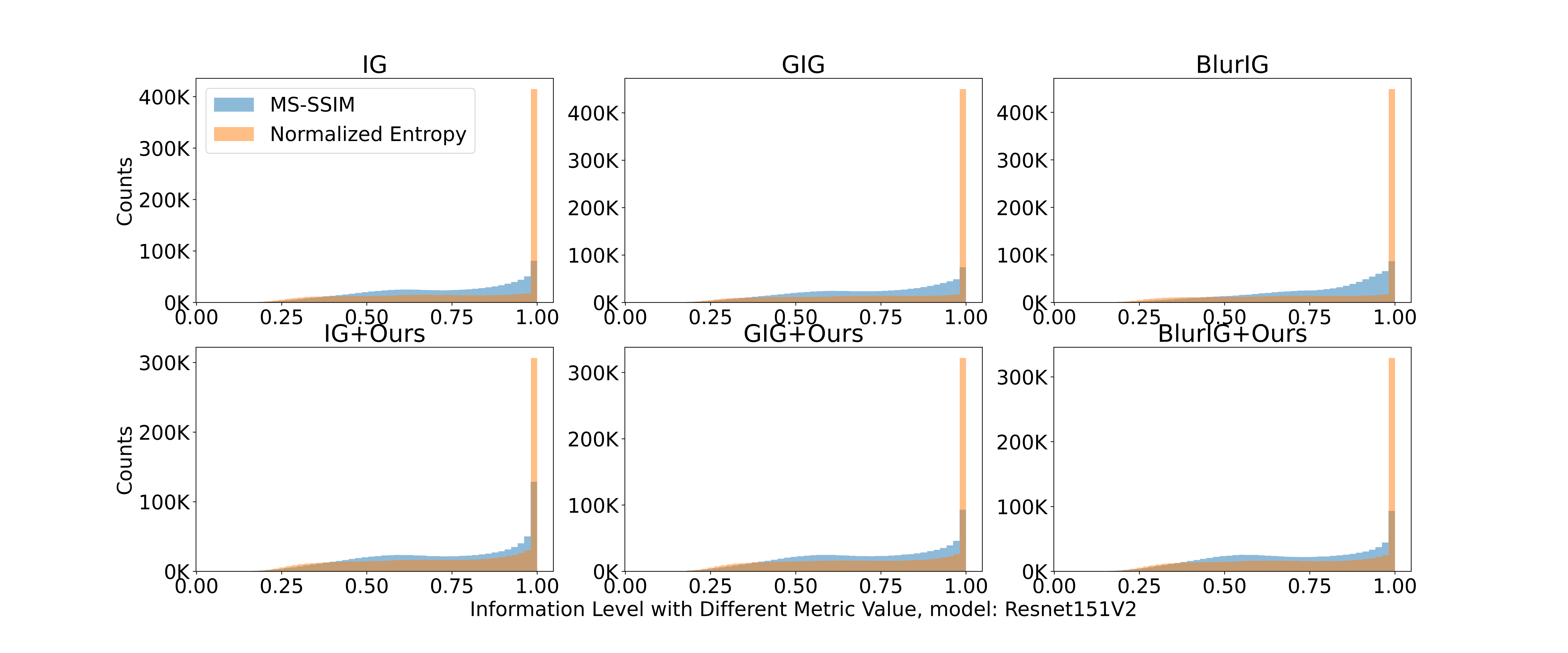}
     \end{subfigure}
\caption{Modified distribution of bokeh images over MS-SSIM and Normalized Entropy  \cite{kapishnikov2019xrai}. Model: \textit{Resnet151V2} 
}
\label{Resnet151V2_f}
\end{figure*}

\begin{figure*}[!t]
     \centering
     \begin{subfigure}[b]{.9\textwidth}
         \centering
         \includegraphics[width=\textwidth]{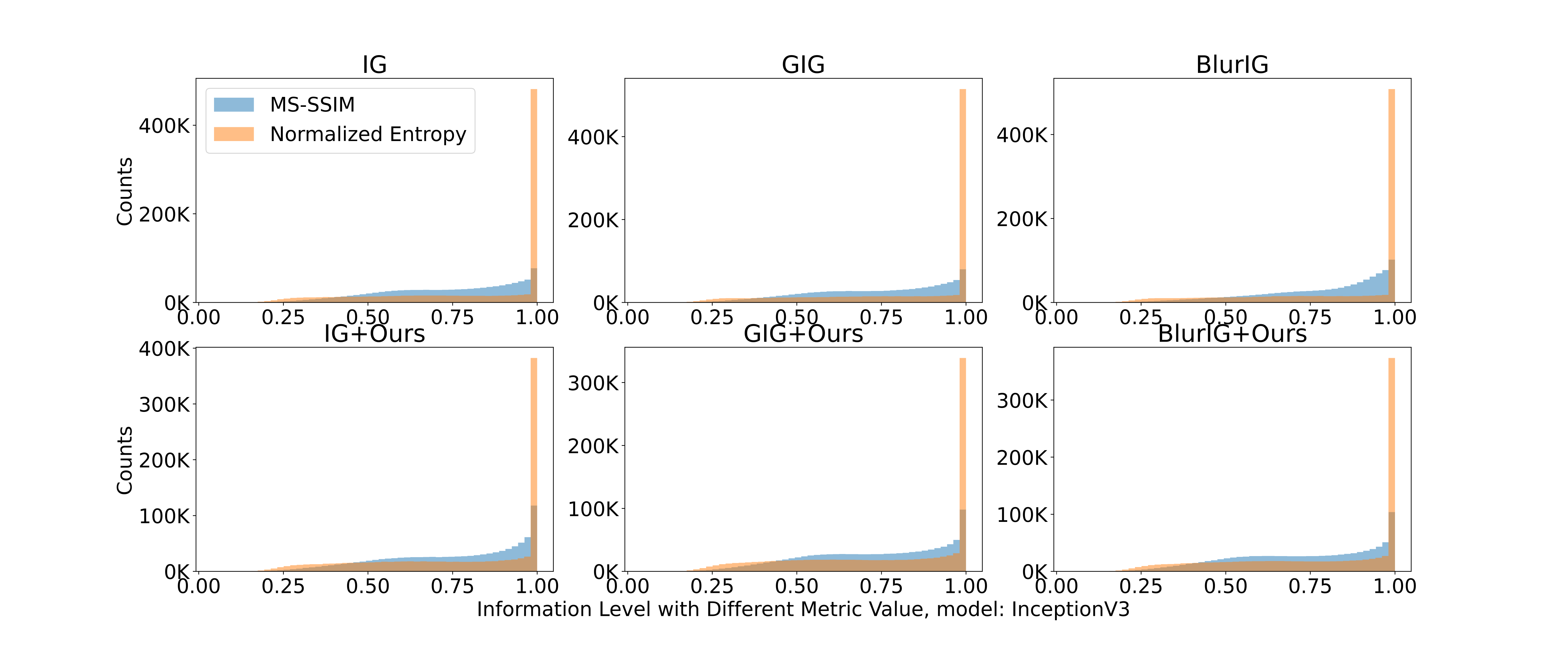}
     \end{subfigure}
\caption{Modified distribution of bokeh images over MS-SSIM and Normalized Entropy  \cite{kapishnikov2019xrai}. Model: \textit{InceptionV3} 
}
\label{InceptionV3_f}
\end{figure*}

\begin{figure*}[!t]
     \centering
     \begin{subfigure}[b]{.9\textwidth}
         \centering
         \includegraphics[width=\textwidth]{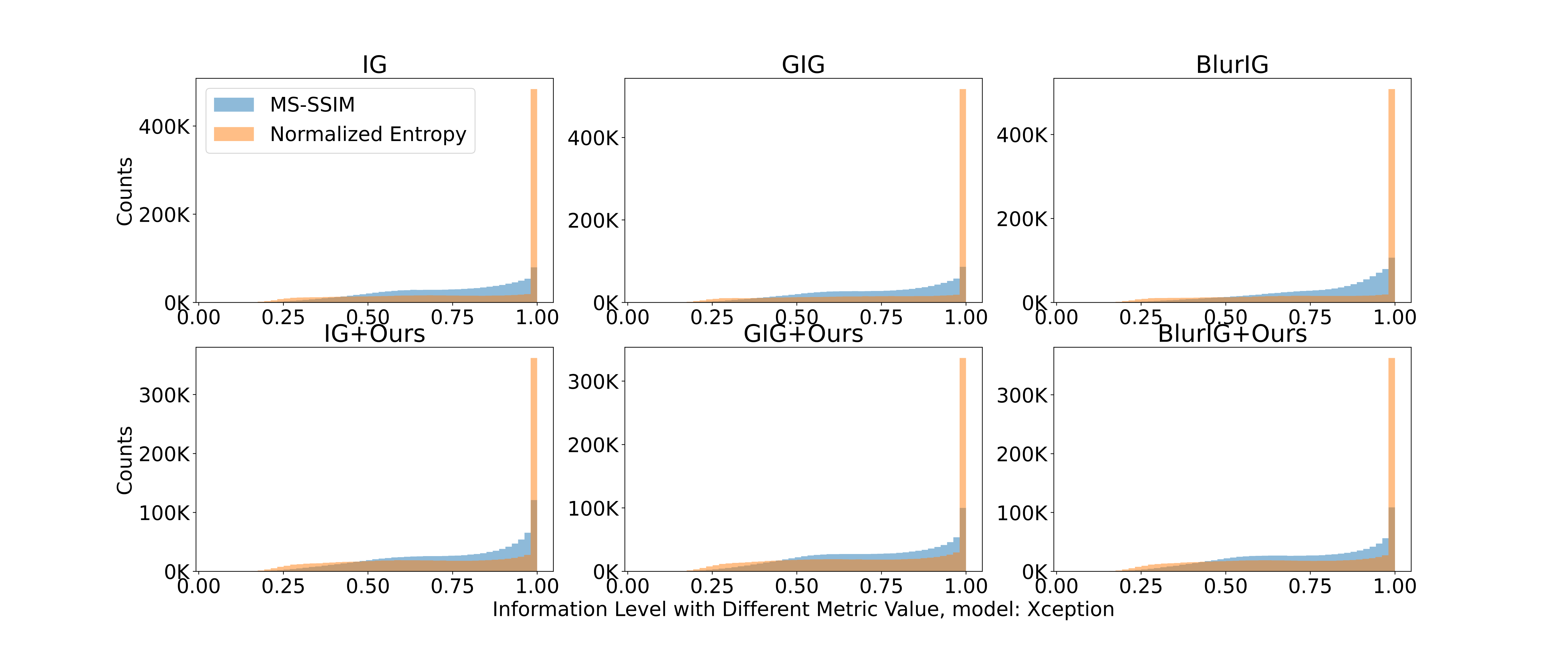}
     \end{subfigure}
\caption{Modified distribution of bokeh images over MS-SSIM and Normalized Entropy  \cite{kapishnikov2019xrai}. Model: \textit{Xception} 
}
\label{Xception_f}
\end{figure*}

\begin{figure*}[!t]
     \centering
     \begin{subfigure}[b]{.9\textwidth}
         \centering
         \includegraphics[width=\textwidth]{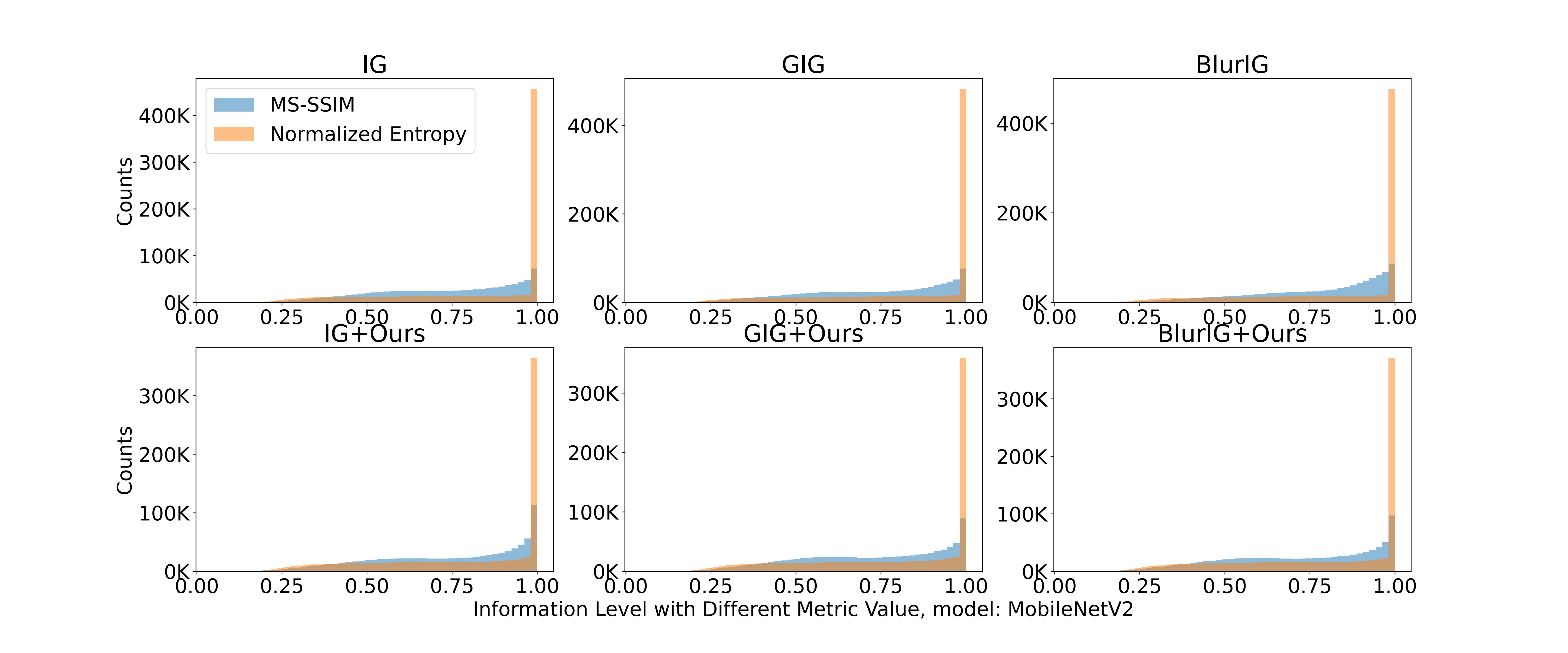}
     \end{subfigure}
\caption{Modified distribution of bokeh images over MS-SSIM and Normalized Entropy  \cite{kapishnikov2019xrai}. Model: \textit{MobileNetV2} 
}
\label{MobileNetV2_f}
\end{figure*}

\begin{figure*}[!t]
     \centering
     \begin{subfigure}[b]{.9\textwidth}
         \centering
         \includegraphics[width=\textwidth]{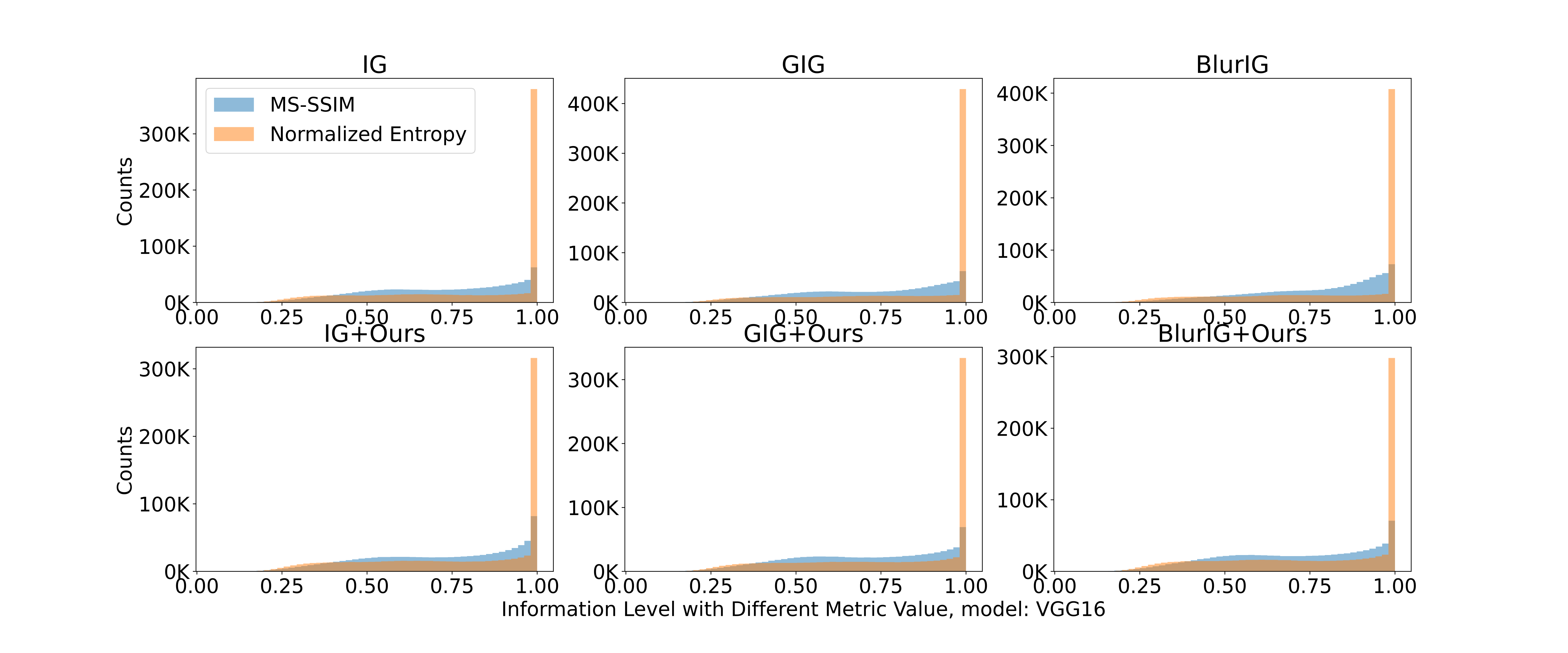}
     \end{subfigure}
\caption{Modified distribution of bokeh images over MS-SSIM and Normalized Entropy  \cite{kapishnikov2019xrai}. Model: \textit{VGG16} 
}
\label{vgg16_f}
\end{figure*}

\begin{figure*}[!t]
     \centering
     \begin{subfigure}[b]{.9\textwidth}
         \centering
         \includegraphics[width=\textwidth]{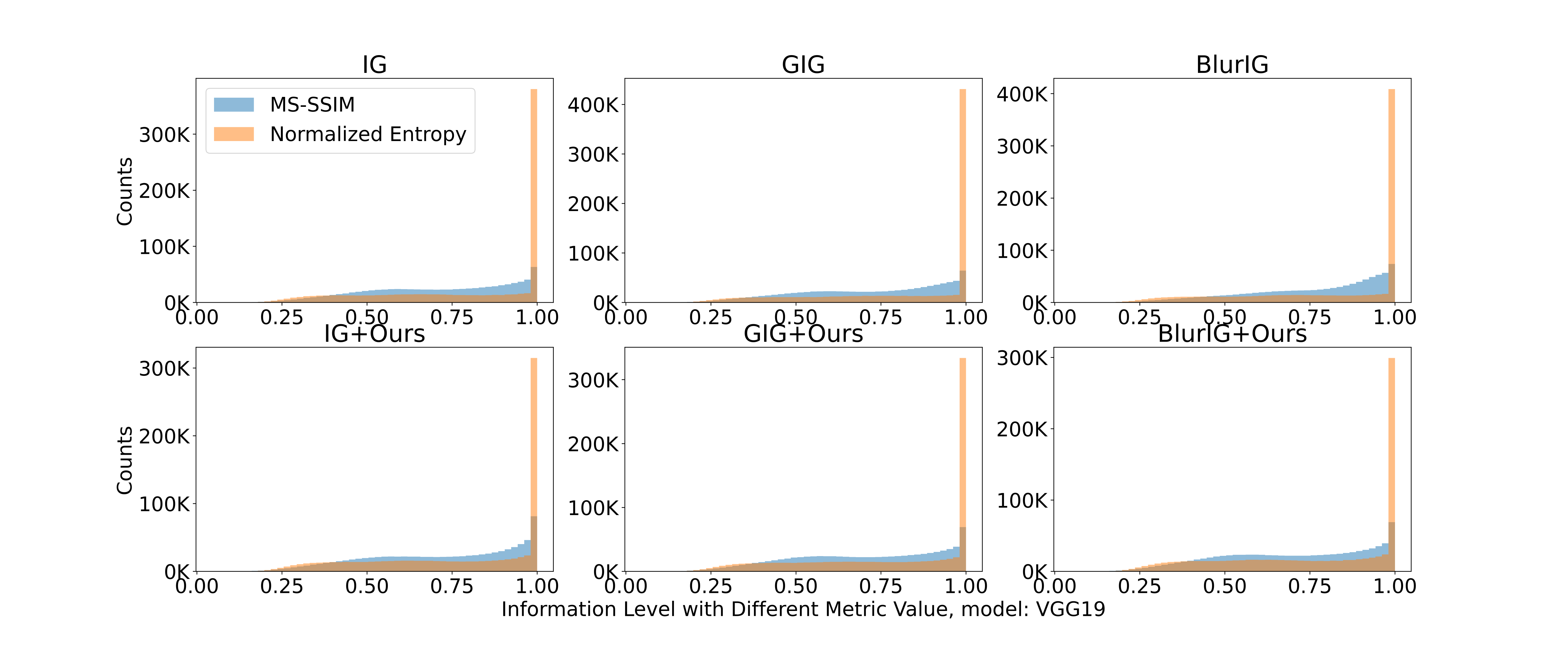}
     \end{subfigure}
\caption{Modified distribution of bokeh images over MS-SSIM and Normalized Entropy  \cite{kapishnikov2019xrai}. Model: \textit{VGG19} 
}
\label{vgg19_f}
\end{figure*}

We present more distribution that compares Normalized Entroy and MS-SSIM in \cref{Dense121_f,Dense201_f,Dense169_f,Resnet101V2_f,Resnet151V2_f,Resnet50V2_f,Xception_f,InceptionV3_f,MobileNetV2_f,vgg16_f,vgg19_f}.

\section{AIC and SIC with XRAI}
\cref{AUC_of_AIC} and \cref{AUC_of_SIC} show the result of AIC and SIC for all methods and its version with XRAI. Similarly, \cref{AUC_of_AIC_SSIM} and \cref{AUC_of_SIC_SSIM} show the result of AIC and SIC with MS-SSIM for all methods and its version with XRAI. 
\begin{table}[!t]
\centering
\resizebox{\columnwidth}{!}{
\begin{tabular}{|c||c|c||c|c||c|c||c|}
\hline
\multicolumn{8}{|c|}{\textbf{AUC of AIC}} \\
\hline
\textbf{Models}&\multicolumn{6}{|c|}{\textbf{IG-based Methods}}&{\textbf{Other}}\\
\cline{1-8}
&IG&+Ours&GIG&+Ours&BlurIG&+Ours&VG\\
\hline

\textit{DenseNet121}&.161&\textbf{.300}&.141&\textbf{.252}&.192&\textbf{.230}&.087\\
\textit{DenseNet169}&.160&\textbf{.288}&.154&\textbf{.254}&.181&\textbf{.216}&.089\\
\textit{DenseNet201}&.185&\textbf{.307}&.182&\textbf{.269}&.213&\textbf{.246}&.110\\
\textit{InceptionV3}&.203&\textbf{.343}&.189&\textbf{.338}&.266&\textbf{.301}&.127\\
\textit{MobileNetV2}&.098&\textbf{.233}&.114&\textbf{.204}&.145&\textbf{.197}&.068\\
\textit{ResNet50V2}&.162&\textbf{.253}&.162&\textbf{.248}&.189&\textbf{.210}&.108\\
\textit{ResNet101V2}&.177&\textbf{.268}&.163&\textbf{.253}&.198&\textbf{.215}&.116\\
\textit{ResNet151V2}&.186&\textbf{.281}&.165&\textbf{.258}&.205&\textbf{.229}&.112\\
\textit{VGG16}&.145&\textbf{.244}&.141&\textbf{.199}&.181&\textbf{.222}&.108\\
\textit{VGG19}&.153&\textbf{.263}&.150&\textbf{.219}&.204&\textbf{.240}&.117\\
\textit{Xception}&.238&\textbf{.404}&.239&\textbf{.381}&.309&\textbf{.355}&.174\\

\hline
&\multicolumn{7}{|c|}{\textbf{With XRAI}} \\
\hline
\textit{DenseNet121}&.438&\textbf{.479}&.460&\textbf{.460}&.437&\textbf{.452}&.434\\
\textit{DenseNet169}&.468&\textbf{.508}&.483&\textbf{.492}&.466&\textbf{.480}&.462\\
\textit{DenseNet201}&.439&\textbf{.476}&.460&\textbf{.468}&.442&\textbf{.461}&.449\\
\textit{InceptionV3}&.477&\textbf{.506}&.472&\textbf{.513}&.479&\textbf{.503}&.496\\
\textit{MobileNetV2}&.407&\textbf{.442}&.437&\textbf{.435}&.410&\textbf{.436}&.424\\
\textit{ResNet50V2}&.402&\textbf{.433}&.428&\textbf{.438}&.409&\textbf{.410}&.417\\
\textit{ResNet101V2}&.415&\textbf{.447}&.433&\textbf{.445}&.416&\textbf{.422}&.424\\
\textit{ResNet151V2}&.410&\textbf{.443}&.421&\textbf{.435}&.406&\textbf{.416}&.412\\
\textit{VGG16}&.393&\textbf{.423}&.422&\textbf{.418}&.402&\textbf{.413}&.396\\
\textit{VGG19}&.386&\textbf{.416}&.417&\textbf{.414}&.396&\textbf{.408}&.393\\
\textit{Xception}&.486&\textbf{.521}&.507&\textbf{.525}&.492&\textbf{.520}&.511\\

\hline
\end{tabular}}
\caption{AUC of AIC}
\label{AUC_of_AIC}
%\vspace{+2mm}
\end{table}

\begin{table}[!t]
\centering
\resizebox{\columnwidth}{!}{
\begin{tabular}{|c||c|c||c|c||c|c||c|}
\hline
\multicolumn{8}{|c|}{\textbf{AUC of SIC}} \\
\hline
\textbf{Models}&\multicolumn{6}{|c|}{\textbf{IG-based Methods}}&{\textbf{Other}}\\
\cline{1-8}
&IG&+Ours&GIG&+Ours&BlurIG&+Ours&VG\\
\hline
\textit{DenseNet121}&.054&\textbf{.228}&.036&\textbf{.157}&.085&\textbf{.134}&.015\\
\textit{DenseNet169}&.052&\textbf{.230}&.045&\textbf{.170}&.083&\textbf{.130}&.016\\
\textit{DenseNet201}&.068&\textbf{.241}&.058&\textbf{.183}&.109&\textbf{.155}&.019\\
\textit{InceptionV3}&.087&\textbf{.294}&.061&\textbf{.286}&.171&\textbf{.232}&.029\\
\textit{MobileNetV2}&.020&\textbf{.145}&.023&\textbf{.111}&.043&\textbf{.103}&.011\\
\textit{ResNet50V2}&.077&\textbf{.210}&.067&\textbf{.201}&.099&\textbf{.158}&.025\\
\textit{ResNet101V2}&.095&\textbf{.231}&.070&\textbf{.201}&.117&\textbf{.165}&.026\\
\textit{ResNet151V2}&.101&\textbf{.249}&.065&\textbf{.212}&.122&\textbf{.177}&.025\\
\textit{VGG16}&.046&\textbf{.166}&.039&\textbf{.104}&.082&\textbf{.141}&.021\\
\textit{VGG19}&.046&\textbf{.177}&.041&\textbf{.115}&.098&\textbf{.151}&.023\\
\textit{Xception}&.119&\textbf{.363}&.107&\textbf{.336}&.218&\textbf{.296}&.054\\
\hline
&\multicolumn{7}{|c|}{\textbf{With XRAI}} \\
\hline
\textit{DenseNet121}&.407&\textbf{.464}&.435&\textbf{.445}&.403&\textbf{.428}&.404\\
\textit{DenseNet169}&.450&\textbf{.496}&.465&\textbf{.475}&.439&\textbf{.458}&.435\\
\textit{DenseNet201}&.427&\textbf{.473}&.449&\textbf{.462}&.419&\textbf{.449}&.432\\
\textit{InceptionV3}&.450&\textbf{.493}&.449&\textbf{.499}&.441&\textbf{.481}&.477\\
\textit{MobileNetV2}&.351&\textbf{.398}&.391&\textbf{.394}&.353&\textbf{.393}&.374\\
\textit{ResNet50V2}&.401&\textbf{.439}&.430&\textbf{.445}&.404&\textbf{.412}&.418\\
\textit{ResNet101V2}&.424&\textbf{.463}&.445&\textbf{.464}&.419&\textbf{.428}&.433\\
\textit{ResNet151V2}&.413&\textbf{.453}&.423&\textbf{.445}&.401&\textbf{.424}&.414\\
\textit{VGG16}&.343&\textbf{.382}&.381&\textbf{.376}&.352&\textbf{.368}&.347\\
\textit{VGG19}&.337&\textbf{.376}&.374&\textbf{.373}&.347&\textbf{.362}&.344\\
\textit{Xception}&.458&\textbf{.502}&.486&\textbf{.508}&.465&\textbf{.503}&.488\\

\hline
\end{tabular}}
\caption{AUC of SIC}
\label{AUC_of_SIC}
%\vspace{+2mm}
\end{table}

\begin{table}[!t]
\centering
\resizebox{\columnwidth}{!}{
\begin{tabular}{|c||c|c||c|c||c|c||c|}
\hline
\multicolumn{8}{|c|}{\textbf{AUC of AIC with MS-SSIM}} \\
\hline
\textbf{Models}&\multicolumn{6}{|c|}{\textbf{IG-based Methods}}&{\textbf{Other}}\\
\cline{1-8}
&IG&+Ours&GIG&+Ours&BlurIG&+Ours&VG\\
\hline
\textit{DenseNet121}&.229&\textbf{.305}&.231&\textbf{.280}&.216&\textbf{.277}&.186\\
\textit{DenseNet169}&.241&\textbf{.314}&.249&\textbf{.297}&.218&\textbf{.289}&.205\\
\textit{DenseNet201}&.254&\textbf{.323}&.262&\textbf{.303}&.237&\textbf{.303}&.216\\
\textit{InceptionV3}&.264&\textbf{.333}&.268&\textbf{.333}&.264&\textbf{.323}&.228\\
\textit{MobileNetV2}&.179&\textbf{.259}&.197&\textbf{.238}&.186&\textbf{.241}&.150\\
\textit{ResNet50V2}&.225&\textbf{.277}&.239&\textbf{.274}&.209&\textbf{.260}&.198\\
\textit{ResNet101V2}&.235&\textbf{.284}&.243&\textbf{.277}&.215&\textbf{.265}&.206\\
\textit{ResNet151V2}&.247&\textbf{.302}&.250&\textbf{.292}&.227&\textbf{.284}&.212\\
\textit{VGG16}&.205&\textbf{.271}&.212&\textbf{.245}&.204&\textbf{.259}&.179\\
\textit{VGG19}&.211&\textbf{.275}&.220&\textbf{.252}&.214&\textbf{.266}&.188\\
\textit{Xception}&.281&\textbf{.362}&.293&\textbf{.356}&.284&\textbf{.345}&.254\\
\hline
&\multicolumn{7}{|c|}{\textbf{With XRAI}} \\
\hline
\textit{DenseNet121}&.342&\textbf{.376}&.360&\textbf{.367}&.336&\textbf{.369}&.351\\
\textit{DenseNet169}&.375&\textbf{.407}&.386&\textbf{.397}&.368&\textbf{.398}&.382\\
\textit{DenseNet201}&.354&\textbf{.388}&.370&\textbf{.380}&.355&\textbf{.387}&.370\\
\textit{InceptionV3}&.357&\textbf{.384}&.355&\textbf{.386}&.348&\textbf{.390}&.373\\
\textit{MobileNetV2}&.310&\textbf{.339}&.333&\textbf{.334}&.310&\textbf{.339}&.329\\
\textit{ResNet50V2}&.302&\textbf{.326}&.320&\textbf{.330}&.302&\textbf{.322}&.317\\
\textit{ResNet101V2}&.316&\textbf{.342}&.329&\textbf{.342}&.312&\textbf{.334}&.327\\
\textit{ResNet151V2}&.314&\textbf{.341}&.321&\textbf{.334}&.308&\textbf{.335}&.321\\
\textit{VGG16}&.314&\textbf{.339}&.334&\textbf{.334}&.319&\textbf{.336}&.319\\
\textit{VGG19}&.309&\textbf{.333}&.330&\textbf{.329}&.315&\textbf{.332}&.315\\
\textit{Xception}&.370&\textbf{.402}&.391&\textbf{.406}&.372&\textbf{.408}&.396\\
\hline
\end{tabular}}
\caption{AUC of SIC with MS-SSIM}
\label{AUC_of_AIC_SSIM}
%\vspace{+2mm}
\end{table}

\begin{table}[!t]
\centering
\resizebox{\columnwidth}{!}{
\begin{tabular}{|c||c|c||c|c||c|c||c|}
\hline
\multicolumn{8}{|c|}{\textbf{AUC of SIC with MS-SSIM}} \\
\hline
\textbf{Models}&\multicolumn{6}{|c|}{\textbf{IG-based Methods}}&{\textbf{Other}}\\
\cline{1-8}
&IG&+Ours&GIG&+Ours&BlurIG&+Ours&VG\\
\hline
\textit{DenseNet121}&.184&\textbf{.263}&.188&\textbf{.239}&.172&\textbf{.236}&.139\\
\textit{DenseNet169}&.205&\textbf{.282}&.214&\textbf{.263}&.182&\textbf{.256}&.166\\
\textit{DenseNet201}&.212&\textbf{.286}&.221&\textbf{.265}&.194&\textbf{.266}&.170\\
\textit{InceptionV3}&.211&\textbf{.287}&.215&\textbf{.285}&.214&\textbf{.276}&.179\\
\textit{MobileNetV2}&.126&\textbf{.204}&.144&\textbf{.187}&.130&\textbf{.188}&.096\\
\textit{ResNet50V2}&.196&\textbf{.254}&.213&\textbf{.250}&.177&\textbf{.236}&.167\\
\textit{ResNet101V2}&.210&\textbf{.265}&.221&\textbf{.256}&.188&\textbf{.244}&.180\\
\textit{ResNet151V2}&.221&\textbf{.282}&.227&\textbf{.270}&.197&\textbf{.261}&.186\\
\textit{VGG16}&.163&\textbf{.234}&.174&\textbf{.210}&.166&\textbf{.224}&.137\\
\textit{VGG19}&.173&\textbf{.240}&.186&\textbf{.219}&.177&\textbf{.233}&.149\\
\textit{Xception}&.223&\textbf{.312}&.233&\textbf{.304}&.229&\textbf{.293}&.194\\
\hline
&\multicolumn{7}{|c|}{\textbf{With XRAI}} \\
\hline
\textit{DenseNet121}&.290&\textbf{.332}&.309&\textbf{.324}&.282&\textbf{.324}&.306\\
\textit{DenseNet169}&.327&\textbf{.364}&.338&\textbf{.356}&.314&\textbf{.353}&.338\\
\textit{DenseNet201}&.311&\textbf{.349}&.326&\textbf{.345}&.301&\textbf{.350}&.333\\
\textit{InceptionV3}&.300&\textbf{.334}&.295&\textbf{.342}&.291&\textbf{.343}&.323\\
\textit{MobileNetV2}&.238&\textbf{.270}&.264&\textbf{.270}&.239&\textbf{.273}&.262\\
\textit{ResNet50V2}&.273&\textbf{.305}&.294&\textbf{.308}&.270&\textbf{.299}&.295\\
\textit{ResNet101V2}&.291&\textbf{.323}&.305&\textbf{.322}&.283&\textbf{.312}&.306\\
\textit{ResNet151V2}&.286&\textbf{.322}&.294&\textbf{.313}&.277&\textbf{.314}&.302\\
\textit{VGG16}&.256&\textbf{.285}&.280&\textbf{.280}&.260&\textbf{.284}&.264\\
\textit{VGG19}&.252&\textbf{.278}&.276&\textbf{.277}&.258&\textbf{.279}&.262\\
\textit{Xception}&.311&\textbf{.345}&.331&\textbf{.352}&.311&\textbf{.353}&.341\\
\hline
\end{tabular}}
\caption{AUC of SIC with MS-SSIM}
\label{AUC_of_SIC_SSIM}
%\vspace{+2mm}
\end{table}
%%%%%%%%% REFERENCES
\newpage
{\small
\bibliographystyle{ieee_fullname}
\bibliography{egbib}
}